\documentclass[10pt]{article} 
\usepackage[accepted]{tmlr}

\usepackage{amsmath,amsfonts,bm}

\def\eqref#1{equation~\ref{#1}}

\def\1{\bm{1}}

\DeclareMathAlphabet{\mathsfit}{\encodingdefault}{\sfdefault}{m}{sl}
\SetMathAlphabet{\mathsfit}{bold}{\encodingdefault}{\sfdefault}{bx}{n}

\usepackage{hyperref}
\usepackage{url}
\usepackage{booktabs}
\usepackage{multirow}
\usepackage{color, colortbl}
\usepackage{graphicx}
\usepackage{comment}

\usepackage[ruled,vlined]{algorithm2e}

\title{Show or Tell? Effectively prompting Vision-Language Models for semantic segmentation}

\author{\name Niccolo Avogaro \email niccolo.avogaro1@ibm.com \\
        IBM Research \\
        ETH Zurich \\ \\
        \name Thomas Frick \email fri@zurich.ibm.com \\
        IBM Research \\ \\
        \name Mattia Rigotti \email mrg@zurich.ibm.com \\
        IBM Research \\
        \\
        \name Andrea Bartezzaghi \email abt@zurich.ibm.com \\
        IBM Research \\ \\
        \name Filip M. Janicki \email fja@zurich.ibm.com \\
        IBM Research \\ \\
        \name Cristiano Malossi \email acm@zurich.ibm.com \\
        IBM Research \\ \\
        \name Konrad Schindler \email schindler@ethz.ch \\
        ETH Zurich \\ \\
        \name Roy Assaf \email roa@zurich.ibm.com \\
        IBM Research \\
}

\def\month{03}  
\def\year{2025} 
\def\openreview{\url{https://openreview.net/forum?id=0yPWtbR3MC}} 

\begin{document}

\maketitle

\begin{abstract}
Large Vision-Language Models (VLMs) are increasingly being regarded as foundation models that can be instructed to solve diverse tasks by prompting without task-specific training.
We examine the seemingly obvious question: \emph{how to effectively prompt VLMs for semantic segmentation}.
To that end, we systematically evaluate the segmentation performance of several recent models guided by either text or visual prompts on the out-of-distribution MESS dataset collection.
We introduce a scalable prompting scheme, \emph{few-shot prompted semantic segmentation}, inspired by open-vocabulary segmentation and few-shot learning.
It turns out that VLMs lag far behind specialist models trained for a specific segmentation task by about 30\% on average on the Intersection-over-Union metric.
Moreover, we find that text prompts and visual prompts are complementary: each one of the two modes fails on many examples that the other one can solve.
Our analysis suggests that being able to anticipate the most effective prompt modality can lead to an 11\% improvement in performance.
Motivated by our findings, we propose PromptMatcher, a remarkably simple training-free baseline that combines both text and visual prompts, achieving state-of-the-art results outperforming the best text-prompted VLM by 2.5\%, and the top visual-prompted VLM by 3.5\% on few-shot prompted semantic segmentation.
\end{abstract}

\section{Introduction}




Large Vision-Language Models (VLMs) have established themselves as the state-of-the-art for cross-modal reasoning that involves images and text, and even as robust backbones for purely visual tasks, benefiting from the wealth of semantic and contextual relations contributed by language modeling. A particular strength of VLMs is the capability to condition image understanding on text inputs, the so-called \emph{Text Prompts} (TP). This enables, for instance, segmentation of a specific object in an image ~\citep{lai2024lisareasoningsegmentationlarge, rasheed2024glammpixelgroundinglarge}, reasoning about relations between objects \citep{you2023ferretrefergroundgranularity, peng2023kosmos2groundingmultimodallarge}, and visual question answering \citep{ beyer2024paligemmaversatile3bvlm, xiao2023florence2advancingunifiedrepresentation}.
Some VLMs also offer conditioning on \emph{Visual Prompts} (VP). Typically these are visual cues like points (suitably embedded coordinates on the image), scribbles or bounding boxes \citep{lai2024lisareasoningsegmentationlarge, rasheed2024glammpixelgroundinglarge}, but it has also been proposed to directly superimpose symbols in pixel space \citep{yang2023setofmarkpromptingunleashesextraordinary}.


We observe that (prompted) VLMs have been studied mainly in two broad settings. The first one could be called \emph{image-driven text generation}, meaning that the system outputs language while visual information is used only on the input side. This setting includes tasks such as image captioning and visual question answering. The second setting can be referred to as \emph{visual grounding}. This setting links language to image regions, helping to enhance the model's spatial reasoning and understanding of how textual descriptions correspond to visual elements in an image. Examples include phrase grounding, where the model is asked to detect the objects mentioned in the text, constraining their spatial relations, and referring expression comprehension, where objects have to be identified based on a periphrasis, thus emphasising contextual relations.

In this work, we focus on the potential of prompting mechanisms to improve image-to-image tasks. Given that large VLMs are increasingly being recognized as foundation models for vision, we ask how to effectively prompt VLMs for semantic segmentation. In other words, our primary interest is not how well the model can parse or generate text about images but rather how accurately it can delineate objects in images.


Since the desired outputs -- segmentation masks -- reside in image space, it is a natural question whether Text Prompts or Visual Prompts are more expedient, and how the two can be combined.
While text prompting has proved successful in guiding image understanding and visual reasoning, we claim that \emph{it is not always sufficient to prompt a VLM with text}, and \emph{visual prompts can in some cases be more suitable, or complementary.} 
Intuitively, a visual example can in certain situations convey information that is much harder, or even impossible, to transmit through text. While the internal mechanisms of large models are notoriously difficult to disentangle and interpret, there is a simple argument in support of visual prompting: The \emph{projection} of the visual world to language is lossy. Even elaborate text descriptions are often ambiguous and can lead to vastly different predictions.



At this point, we must highlight a subtle but important difference that is sometimes overlooked: text prompts are normally understood as generic statements that can be defined once and then applied across many images, like ``segment all cats''.
In contrast, visual prompts are predominantly understood as image-specific, like for instance a scribble to denote the cat in a particular image.
In this interpretation, visual prompting requires user input for every new sample and is not scalable. Instead, we advocate for a form of visual prompting that incurs only a constant overhead for arbitrarily large test sets: The user annotates instances of their desired target class on a small number of images, then that fixed set of examples serves as the prompt for any new test image, and no further interaction is expected. 
%
We refer to this setup as \emph{few-shot prompted semantic segmentation} (FPSS). Unlike traditional few-shot learning, which also uses a small set of annotated examples but requires fine-tuning the model, FPSS operates through prompting rather than training. It is also related to \emph{training-free open-vocabulary segmentation}, where a frozen model is adapted to new classes without retraining, though typically in a zero-shot context rather than using a few-shot approach.



When evaluating under the FPSS protocol, we find that VLMs are not (yet) generic, \emph{foundational} representations. They still trail domain-specific segmentation models by about 30\% on average in Intersection-over-Union (IoU) score on the dataset used in this work.
Furthermore, we find that text prompts perform better \emph{on average}, but that visual prompts are able to address tasks that are exceptionally difficult for text prompted models. Unsurprisingly, the two prompting modes are to some degree complementary: in hard scenarios, e.g.~medical imaging, VP can solve many instances that TP cannot, and vice versa.


Motivated by these findings, we construct a simple baseline for combined text and visual guidance while still maintaining a training-free, prompting-only setup. Prompting with both text and vision indeed improves the performance by a significant 2.5\% compared to only text (respectively, 3.5\% compared to only vision). 


Summarizing our contributions:

\begin{itemize}
    \item We design a novel benchmarking task to probe the performance of VLMs as semantic segmentation engines.
    \item We show that even the latest models remain far below custom models trained for a specific task and data domain. In other words, we are still far from \emph{foundational} VLMs, i.e. having a homogenized VLM capable of solving multiple tasks in a zero-shot manner.
    \item We show that text and visual prompting complement each other, and that being able to anticipate the most effective prompt modality can lead to a 11\% improvement in performance.
    \item We propose a simple training-free framework to capitalize on the complementary strengths of text and visual prompts and achieve state-of-the-art on the MESS dataset collection \cite{blumenstiel2023messmultidomainevaluationzeroshot}.
\end{itemize}


\section{Task formulation}
\label{sec:task}
\begin{figure}[t]
    \centering
    \includegraphics[width=1\textwidth]{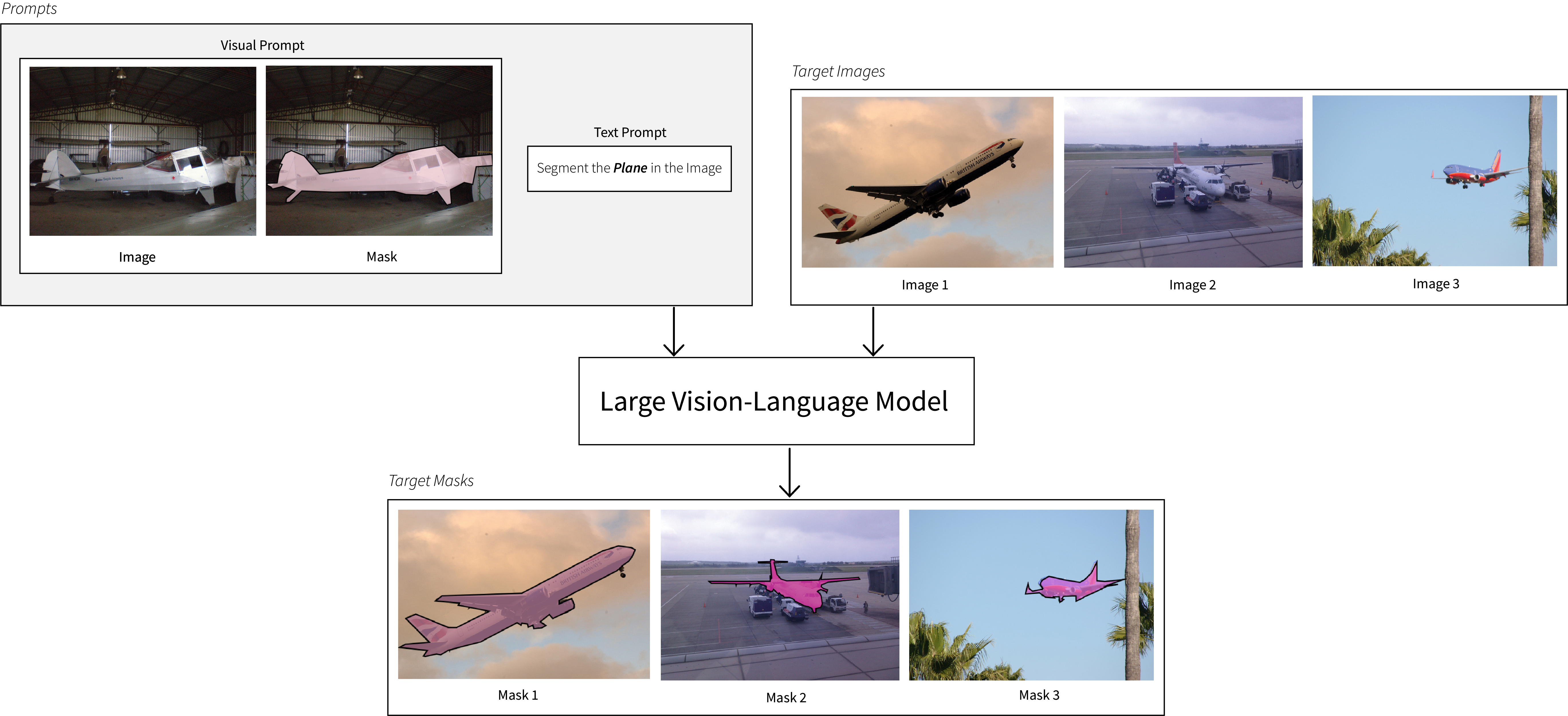} 
    \caption{The FPSS task involves providing a VLM with visual (image + corresponding mask) and text prompt. The goal is for the model to make predictions on new target images.}
    \label{fig:task} 
\end{figure}








The goal of our paper is to evaluate to which extent (training-free) prompting of generalist VLMs can replace specialist models for semantic segmentation.
%
It is obvious that some form of prompt is always required to let a VLM know what to segment, but it is much less obvious what the most suitable prompt is. Here, we limit ourselves to the two most popular ones, text and visual prompts. 

As an example, let us assume we want to segment airplanes. A natural way to instruct the model is with one or a few text prompts, like ``segment all airplanes". 
Note that, due to the compositional nature of language, there is no clear definition of how many prompts we are effectively using since two or more prompts can be merged into one, as in ``segment airplanes and similar flying machines". 
In normal text prompting, the same prompt is then applied to all input images. 
FPSS translates that one-off prompting scenario to the visual domain: the user supplies the system with at most $K$ reference images of airplanes, along with their segmentation masks or other annotations (e.g., a set of points within the mask). Based on that input, the system shall segment airplanes in any number of unseen target images. 
Note that this mode of interaction makes it possible to communicate about visual concepts whose category name is not known to the model, just like a child can say ``I want this" before learning the word ``cookie".

We now highlight how the FPSS task differs from classical few-shot segmentation (FSS), open-vocabulary semantic segmentation (OVSS), and referring segmentation. 
In FPSS, the base model is a foundation model trained on internet-scale datasets, and the task involves only prompting the model at test time. This is done by providing a text prompt in the form of a sentence, and a visual prompt in the form of a reference image.
In contrast, FSS is solving a semantic segmentation problem using solely visual examples. Typically, it involves training a model on a meta-task with a defined set of classes, then evaluating the model on similar domains with different test classes. Some FSS techniques include test-time training. 
FPSS differs from OVSS by supporting visual prompts in addition to text. Additionally, FPSS allows free-form language inputs.
As for the referring segmentation task, that involves generating a segmentation mask for a particular object specified by a natural language expression (text prompt), which requires precise alignment between the linguistic description and the visual content. In contrast, FPSS also makes use of visual prompts applied at the semantic level and not targeting specific instances, therefore it does not cater to the fine-grained, instance-specific requirements of referring segmentation, where accurately locating and isolating a single object is crucial.


To adapt existing datasets for FPSS, we can employ two distinct approaches. We can retrofit OVSS datasets like MESS by generating visual prompts from instances with the same semantic class. For FSS datasets (COCO-20i and PASCAL-5i) we can generate a textual prompt by utilizing the class names. On the other hand, referring segmentation datasets (e.g., refCOCO, refCOCO+) present a fundamental limitation for FPSS adaptation, as FPSS works at the semantic level, rather than the instance level due to the usage of visual prompts.

Beyond the research questions on how the two prompting modes compare and when one or the other is more successful, prompting in the FPSS setting is relevant in several real application scenarios as digitalization and AI permeate society. For instance, an engineer may have to instruct an inspection system to examine a new item, or a biologist may want to screen a legacy image collection for a newly discovered species. In both scenarios, users may prefer to provide only a few text or visual prompts to the system, expecting the task to be automatically applied to an entire dataset of unseen images.

\section{Analysis}
\label{sec:ana}

In this section, we outline the evaluation framework and specify the models considered used in our FPSS setup, specifically under the one-shot regime. We select a range of key text prompted, respectively visually prompted models, and assess their effectiveness in performing segmentation when provided with the corresponding prompt modality. We then present and discuss the results, providing a detailed analysis of the performance differences across modalities, highlighting strengths and limitations. 

\subsection{Evaluation protocol}

There are many models capable of performing segmentation guided by text prompts, mainly falling into two categories: open-vocabulary segmentation models \citep{cho2024catsegcostaggregationopenvocabulary}, \citep{hajimiri2024payattentionneighbourstrainingfree} and vision-language models (VLMs) \citep{lai2024lisareasoningsegmentationlarge, beyer2024paligemmaversatile3bvlm}. Both types of models leverage textual input to guide segmentation, with open-vocabulary models focusing specifically on identifying objects beyond a fixed set of categories, while VLMs, with their broader multi-modal capabilities, can also be adapted for segmentation tasks.
Similarly, we identify two categories of models that can be prompted visually: models specifically trained with visual prompts \citep{li2023visualincontextprompting, zou2023segment} and training-free frameworks that leverage existing segmentation models along with matching algorithms \citep{liu2024matchersegmentshotusing, matcher_vp}. 
In contrast, very few models have been presented that can be guided with both text and visual prompts \citep{zou2023segment}

For open-vocabulary segmentation models, we consider the fully-supervised approach CAT-Seg \citep{cho2024catsegcostaggregationopenvocabulary}, the state of the art on the MESS dataset and the training-free approach NACLIP \cite{hajimiri2024payattentionneighbourstrainingfree}. In particular, we use CAT-Seg with the \textit{CLIP ViT-L/14} backbone, and NACLIP with the standard \textit{CLIP ViT-B} configuration. We also include SEEM \citep{zou2023segment}, specifically the SEEM \textit{Davit-Large} implementation. This is the only available model to accept TPs and VPs simultaneously, although in this section, we only use them separately. Combined prompting with SEEM is discussed in Section~\ref{sec:method}.

As VLM baselines, we include the decoder-free Florence-2 \citep{xiao2023florence2advancingunifiedrepresentation}, specifically the segmentation branch of the large, fine-tuned model, where we clip the generated sequence length to 1024 for computational reasons; and PALI-Gemma \citep{beyer2024paligemmaversatile3bvlm}, a small but effective architecture that uses a VQVAE decoder \cite{oord2018neuraldiscreterepresentationlearning}. We use the standard \textit{224-mix} implementation.
We also evaluate the recent LISA \citep{lai2024lisareasoningsegmentationlarge}, in particular, the \textit{LISA-13B-llama2-v1} version. 
LISA integrates a Multi-modal Large Language Model (LLaVA \cite{liu2023llava}) with a CLIP vision backbone and SAM. The model introduces a special <SEG> token to the LLM's vocabulary, employing an embedding-as-mask paradigm, where the hidden state corresponding to the <SEG> token is used by a fine-tuned SAM mask-decoder to generate segmentation masks. To keep the evaluation focused, and taking into account computational resource limitations, we regard LISA as proxy for its descendants GLAMM \citep{rasheed2024glammpixelgroundinglarge} and SESAME \citep{wu2023seesaysegmentteaching}, which might offer marginal improvements.
Our choice of VLMs is primarily informed by their performance in terms of referring segmentation on the RefCOCO, RefCOCO+, and RefCOCOg datasets \citep{Kazemzadeh2014ReferItGameRT, mao2016generationcomprehensionunambiguousobject}, a task that is closely related to our FPSS task. In all cases, we opt for greedy LLM decoding. 

When considering models that are specifically trained with visual prompts, we once more pick SEEM \citep{zou2023segment}, using the same implementation as described for the text prompting setting, as well as DINOv \citep{li2023visualincontextprompting}, using its Swin-L variant. 
%
Regarding visually prompted training-free frameworks, we choose Matcher \citep{liu2024matchersegmentshotusing} motivated by its performance on COCO-20i, and its follow-up work SoftMatcher \citep{matcher_vp} mainly for its computational efficiency, both of which leverage pre-trained foundation models, namely Segment Anything \citep[SAM,][]{kirillov2023segment} and DINOv2 \citep{oquab2024dinov2learningrobustvisual}, in combination with traditional matching algorithms to provide image-prompted segmentation capabilities.
Furthermore, we modify the SoftMatcher framework to obtain an improved version, which we call SoftMatcher+. The resulting algorithm follows a five-step pipeline: i) features are extracted from both reference and target images using the vision foundation model. ii) These features are then used for probabilistic feature matching to produce a set of probable object locations (points) in the target images. iii) The point prompts are fed to the SAM model to produce a set of mask proposals. iv) Each mask proposal is tested for consistency with the reference and rejected masks are discarded. v) Masks that have passed the rejection step are merged via binary union to obtain a single, final output mask.  SoftMatcher+ utilizes AM-RADIO \citep{ranzinger2024amradioagglomerativevisionfoundation} as its backbone instead of DINOv2, leveraging the excellent abilities of AM-RADIO features (distilled from several large models including CLIP, DINOv2, and SAM) in terms of matching, pixel-level localization, and vision-language connections. For all these training-free methods, we make use of the ViT-L versions of the models (DINOv2, SAM, AM-RADIO) and tune their hyper-parameters on COCO-20i.

Regarding text prompts, we proceed as follows: for open-vocabulary segmentation models that accept only a class name as input, we use class names based on the dataset specifications. For VLMs with advanced language abilities, we embed the class name in the sentence ``Segment all the instances of class \texttt{class\_name} in the image''. 
As visual prompts, we sample one single image of the target class from the dataset itself, together with its ground truth segmentation mask. The minimal setup with a single prompt image, respectively an elementary text prompt, is a challenging and particularly user-friendly scenario. Picking the prompt image from the same dataset corresponds to the realistic scenario where the user creates the prompt on images acquired in their application setting, with similar imaging conditions and class definitions as the test data. To minimise biases due to the choice of prompt image, we sample a different prompt image for each prediction.

We point out that both text prompts and visual prompts can be refined by prompt engineering. 
Various techniques have been proposed, ranging from single prompt optimization \citep{zhou2023leasttomostpromptingenablescomplex} through prompt ensembling \citep{wang2023selfconsistencyimproveschainthought} to multi-step reasoning \citep{wei2023chainofthoughtpromptingelicitsreasoning, yao2023treethoughtsdeliberateproblem, zhang2024multimodalchainofthoughtreasoninglanguage}.
While prompt engineering can make a substantial difference, it has become an art in itself and, in fact, an entry barrier for inexperienced users. It goes beyond the scope of the present work but may be an interesting avenue for future research.


We also consciously refrain from any fine-tuning. Often, even large models are fine-tuned for specific tasks, which can significantly improve their performance. In our view, this strategy is somewhat misaligned with the definition and purpose of a "foundation model", which should ideally be usable with minimal intervention. Once the hardware, data, and expertise for fine-tuning are required, there is arguably little qualitative difference from the well-established practice of training a dedicated model starting from pre-trained weights (e.g., from ImageNet).

As a testbed for our experiments, we use the MESS dataset collection \citep{blumenstiel2023messmultidomainevaluationzeroshot}. It consists of 22 different segmentation datasets that span a wide variety of application domains and image characteristics. The datasets are grouped into five broad domains, \emph{General} (6 datasets), \emph{Earth} (5), \emph{Medical} (4), \emph{Engineering} (4) and \emph{Agriculture} (3) as detailed in Table~\ref{tab:mess-groups}. The MESS dataset collection was deliberately designed as a challenging benchmark for open-vocabulary models and is an ideal choice for evaluating foundation models. This is because its constituent datasets span a wide range of domains and target categories with different image characteristics, many of which differ significantly from the dominant benchmark datasets and the generic scraped internet data that are typically used to train VLMs.
Moreover, MESS comes with strong baselines generated with per-dataset, domain-specific semantic segmentation models. 
%


We intentionally refrain from conducting evaluations on widely-used benchmarks such as COCO \citep{lin2015microsoftcococommonobjects}, Pascal VOC \citep{pascal-voc-2012}, and LVIS \citep{gupta2019lvisdatasetlargevocabulary}. These datasets are likely included, either partially or fully, in the data used to train popular foundation models, and thus risk information leakage. To ensure a rigorous and unbiased assessment, we prioritize evaluations on the out-of-domain datasets in the MESS collection, thereby providing a more robust measure of the models' generalization capabilities and their ability to cope with in unseen conditions. We ground this statement in Section~\ref{sec:indomain}

For clarity of presentation, we always show average numbers for the five broad domains covered by MESS. The detailed dataset composition is provided in Appendix~\ref{sec:appendix_mess}. We compare our evaluation of FM capabilities to the \textit{supervised baselines} which we sourced directly from the MESS paper. As detailed in their Supplementary Material, Section B.1, the authors compiled these baselines per dataset using the best-performing supervised models trained in-domain. In datasets where such models were unavailable, they trained a supervised model themselves.

The evaluations were run on a single A100 with 40GB of memory, which takes $\approx$14 hours for one complete run with the largest model (LISA-13B). Open-vocabulary segmentation models are faster, completing one evaluation cycle in 9 hours, while Florence-2 is the slowest, taking almost 24 hours. 
Visually prompted models are substantially lighter (up to 1.2B parameters) than their text prompted counterparts (up to 13B parameters), and while Matcher is very slow (22 hours), SoftMatcher+ takes around 5 hours for an evaluation cycle.

\subsection{Results}
\label{sec:text_prompting}

\begin{table}[t]
\centering
\begin{tabular}{l|ccccc|c}
\toprule
& General & Earth & Medical & Engineering & Agriculture & Average \\	
\midrule
SEEM text   & 35.9  & 36.8  & 28.9  & 13.9  & 44.5  & 32.0 \\
CAT-Seg     & 33.9  & 36.9  & \textbf{45.7}  & \textbf{48.4}  & 24.5  & 37.9 \\
Florence    & 14.0  & 13.9  & 13.1  & \textcolor{white}{0}7.3  & \textcolor{white}{0}7.6  & 11.2 \\
PALI-Gemma  & 35.3  & 29.1  & 28.4  & \textcolor{white}{0}7.2  & 40.0  & 28.0 \\
NACLIP & 36.1  & 41.2  & 22.7  & \textcolor{white}{0}6.8  & 22.3  & 25.8 \\
LISA        & \textbf{57.0}  & \textbf{47.6}  & 31.6  & 12.7  & \textbf{63.9}  & \textbf{42.6} \\
\midrule
SEEM Vision      & \textcolor{white}{0}9.6 & 16.8  & 20.5  & \textcolor{white}{0}6.9  & 21.7  & 15.1 \\
DINOv           & 37.4  & 28.0  & 24.2  & \textcolor{white}{0}8.3   & 59.1  & 31.4 \\
Matcher         & 43.2 & 31.2 & 26.0 & 12.4 & 54.9 & 33.5 \\
SoftMatcher     & 48.0  & 34.0  & 31.5 & 18.8  & 59.8  & 38.4 \\
SoftMatcher+    & \textbf{53.0}  & \textbf{36.2}  & \textbf{30.4}  & \textbf{28.7}  & \textbf{60.7}  & \textbf{41.8} \\

\midrule
Supervised  & 55.2  & 71.4  & 82.6  & 89.4  & 62.8  & 72.3 \\
\bottomrule
\end{tabular}
\caption{Evaluation results on the MESS dataset. The table presents performance metrics for text-prompted models (first block), visual-prompted models (second block), and supervised baselines (last row).}

\label{table:performance_text_only}
\end{table}

Table~\ref{table:performance_text_only} showcases the results under the FPSS evaluation scenario on the MESS dataset. Notably, we see that all the evaluated promptable models still trail domain-specific segmentation models by about 30\% IoU on average. In the second block of Table~\ref{table:performance_text_only}, we see that among text prompted models, while NACLIP underperforms - a behavior that might be attributed to its training-free nature - CAT-Seg and SEEM remain competitive baselines when compared to the VLM approaches. In fact, with the exception of LISA, the LLM-based methods underperform relative to these baselines. We hypothesise that this performance is attributed to mainly two factors. First, the detokenization procedure employed by these models could lack the granularity required for dense tasks.
Second, the training data for these models encompasses a broad range of image reasoning tasks beyond segmentation, including visual question answering, object detection, and visual grounding. This diversity in training, while beneficial for general-purpose applications, may dilute the models' effectiveness on segmentation tasks.

Moreover, LISA emerges as the front-runner, with an average IoU of $42.6\%$, around $4.5$ IoU points higher than the second-best performing model CAT-Seg. This is likely due to LISA's specialized foundation model decoder and to its extensive training regimen on the large segmentation dataset SA-1B \citep{kirillov2023segment}, which is then further aligned with segmentation-specific datasets such as RefCOCO or ADE20K \citep{zhou2018semanticunderstandingscenesade20k}.
More interestingly, comparing LISA with domain-specific models trained on individual datasets yields an important finding: we find that in some cases, LISA outperforms the baseline on generalist tasks, surpassing specialized segmentation models optimized for in-domain performance. 
We hypothesize that this performance gain is due to LISA being trained on a larger, more diverse, and generalist dataset that closely aligns with the classes found in some general MESS datasets such as DRAM, ATLANTIS, SUIM, etc. The supervised baseline, by contrast, is trained on a smaller, task-specific dataset, which limits its exposure to a wide range of examples. LISA's broader training enables it to learn more robust features, leading to better performance on generalist datasets.

However, it is also crucial to note that LISA's performance significantly decreases in more technical domains, such as engineering and medical applications. In these specialized areas, it is surpassed by the open-vocabulary segmentation models, particularly CAT-SEG, and by domain-specific models. This performance gap in technical domains suggests potential for improvement.

The second block of Table~\ref{table:performance_text_only} presents the results of the visual prompted models. We see that these models underperform on average compared to their text prompted counterparts. For instance, the performance of SEEM Vision is significantly inferior to SEEM Text. And while SoftMatcher narrows this performance gap, SoftMatcher+ demonstrates even better results, nearly reaching LISA's performance level.
In particular, we highlight that SoftMatcher+ shows superior performance compared to LISA on the technical domains. We attribute this improvement to the nature of image examples, which more precisely and effectively capture the user's interests with better precision and varying levels of detail. 

\section{Show or tell?}
\label{sec:showtell}

Our findings in Section~\ref{sec:text_prompting} suggest that visual prompting and text prompting behave differently when it comes to different target domains.
To gain deeper insights into this performance disparity, we conduct a more thorough examination of the top-performing models from each category.
This comparative analysis helps us elucidate the factors underlying the performance differences between visual and text-based prompting.

\subsection{Oracle ensembling of text and visual prompts}

\begin{table}[t]
\centering
\begin{tabular}{l|ccccc|c}
\toprule
& General & Earth & Medical & Engineering & Agriculture & Average \\
\midrule
SoftMatcher+        & 53.0 & 36.2  & 30.4  & 28.7  & 60.7  & 41.8 \\
LISA                & 57.0  & 47.7  & 31.7  & 12.8  & 64.0  & 42.6 \\
Oracle Ensemble     & 60.9  & 47.8  & 40.4  & 28.7 & 65.4 & 48.6 \\
Oracle Ensemble+    & \textbf{67.3}  & \textbf{51.8}  & \textbf{46.2}  & \textbf{32.5}  & \textbf{71.4} &\textbf{53.8} \\
\midrule
Supervised          & 55.3  & 71.4  & 82.6  & 89.5  & 62.8  & 72.3 \\
\bottomrule
\end{tabular}
\caption{Oracle ensemble methods compared to the best performing text and visual prompt models, and to the supervised baseline.}
\label{table:task}
\end{table}

A natural starting point for characterizing the differences between visual and text prompting is to determine by how much the segmentation performance improves by choosing the best prompting modality \emph{within each target domain}.
Regarding the MESS datasets, this can be easily quantified by choosing the better performing method between VP (SoftMatcher+) and TP (LISA) for each dataset, obtaining what we call the \emph{Oracle Ensemble}.
Table~\ref{table:task} shows that being able to choose optimally between using visual or text prompts (per dataset) brings a boost to the overall performance by 6\% compared to LISA.

Motivated by this, we add more granularity to this analysis and investigate the performance upper bound that we could reach by selecting the best prompting technique (VP or TP) on a \emph{per-image} basis, as opposed to \emph{per-dataset} (as in Oracle Ensemble).
We denote the resulting upper bound with \emph{Oracle Ensemble+} and note in Table \ref{table:task} its remarkable performance of 53.8\%, corresponding to an 11\% jump over pure text prompting with LISA.

The simple baselines given by these Oracle Ensembles show the potential advantages of using visual prompts in conjunction with conventional text prompts.
In addition, given their simplicity, they highlight the possibility that more advanced models, with access to both modalities, could achieve even greater performance when coupled with a smart integration of both sources. 
This motivates us to seek ways to leverage visual prompting in text prompted VLMs.



To optimally leverage visual prompts we first investigate the source of its relative advantage over text prompts. 
Looking at IoU differences on a per-class basis and ranking them based on the absolute difference as shown in Table~\ref{tab:performance_difference}, we uncover a striking trend. The top 10 values all favor VP, with some classes showing a remarkable performance advantage of up to 80\%. This substantial disparity underscores the significant superiority of visual prompting over text prompting for certain classes, suggesting that visual cues provide a more effective means of guiding the model's segmentation process in these instances.

This analysis across different class names suggests that the shortcomings of text prompted models are not primarily due to an inability to segment specific objects, but rather stem from the nature of the prompts themselves. The classes where LISA performs poorly fall into two main categories: ambiguous descriptions such as \textit{Upper clothes} and highly specific, uncommon class names such as \textit{ Worm-eating warbler} or \textit{Fjord}. These findings suggest that the model's difficulties arise from interpreting vague or extremely niche text prompts, rather than from fundamental limitations of its latent image encoding.

To better understand the performance discrepancies, we visually inspect samples from the first four categories, i.e., samples representing the most divergent IoU scores per class. The qualitative results can be seen in the first four columns of the Figure~\ref{fig:Worm-eating warbler}.
On the first sample of class \emph{Worm-eating warbler}, the model clearly struggles to interpret the user's request, failing to connect the specific subclass to the broader \textit{bird} category, despite the relative segmentation-friendly image content.
On the second sample, the model produces only noise at the top of the image, demonstrating a complete failure to identify the requested class of Rape (referring to the Rapeseed plant). 
The third sample reveals the model's confusion between segmenting the mountain portion of the fjord and the fjord itself, resulting in an inaccurate segmentation of the mountain.
In the fourth example, LISA exhibits hallucination, segmenting an unrelated object when asked to segment the class \emph{Date}. 

\subsection{Ambiguity of text prompting}
\label{sec:text_ambiguity}

\begin{table}[t]
\centering
\begin{tabular}{l|ccc}
\toprule
Class name & IoU TP	& IoU VP & IoU Difference \\
\midrule
Worm-eating Warbler & \textcolor{white}{0}1.4 & 82.2 & 80.8 \\
Rape & 19.2 & 80.0 & 60.8  \\
Fjord & 24.1 & 81.2 & 57.0  \\
Date & \textcolor{white}{0}0.1 & 52.0 & 51.9  \\
Hair & 18.8 & 62.1 & 43.2 \\
Upper clothes & 16.0 & 58.2 & 42.2  \\
Tea & 29.9 & 70.5 & 40.6  \\
Soy & 37.2 & 77.2 & 40.0  \\
Cashew & 27.7 & 66.9 & 39.1 \\
Kiwi & 37.3 & 76.3 & 39.0 \\
\bottomrule
\end{tabular}
\caption{Top 10 classes with the highest IoU difference between the text and visual prompt models.}

\label{tab:performance_difference}
\end{table}

\begin{figure}[t]
    \centering
    \includegraphics[width=0.9\textwidth]{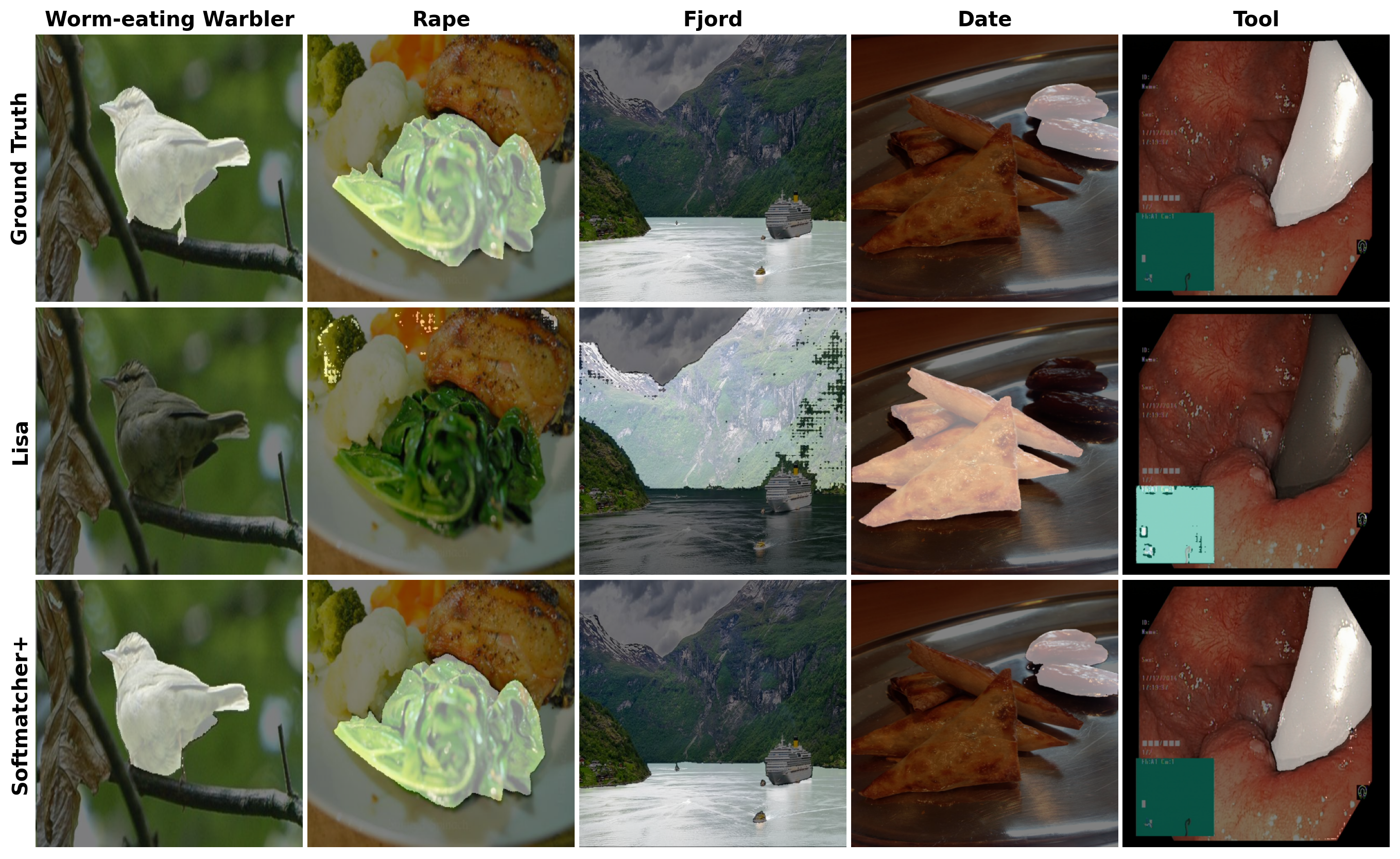} 
    \caption{Qualitative analysis of the results of LISA and SoftMatcher+ compared to ground truth. The first four columns display images selected according to biggest difference of IoU between VP and TP as per Table~{\ref{tab:performance_difference}}. The last column displays the \emph{Tool} class.}
    \label{fig:Worm-eating warbler} 
\end{figure}

The visual inspection of the top samples in terms of performance difference between TP and VP suggests that the discrepancies can be attributed to two main linguistic challenges: ambiguity from polysemous or homonymous words and the use of highly specialized or uncommon terms.

These issues are closely related to the inherent complexities of language, which complicate the ability of text prompted systems to accurately interpret visual tasks.
The interplay between ambiguity and specificity in language is inherent on how it was formed \citep{Riemer_1950} and it is widely known to be an issue in the computational semantics literature, hindering the algorithmic performance \citep{church-patil-1982-coping, Manning}.
The trade-off between the usage of ambiguous words and ones that are specific, unusual, or difficult to pronounce serves a crucial role in our ability to convey complex thoughts and adapt to diverse communicative contexts \citep{Wasow2015-WASAAI}. 


Our hypothesis that language ambiguity can be a considerable weakness for visual prompting is supported by further experiments on the MESS FoodSeg103 dataset.
Here, we see a significant performance gap of 13\% of IoU between Oracle Ensembling (which in this case refers to LISA) and Oracle Ensembling+. This can be attributed to the linguistic challenges previously discussed. 
FoodSeg103 encompasses a diverse set of food categories, many of which are either ambiguous or highly specific, making them challenging to distinguish through text description. On the other hand, these foods often appear visually similar. Additional examples are provided in Appendix~\ref{section:appendix_qualitative}.

Similarly, the Kvasir-Inst.\ dataset shows a notable discrepancy, particularly for the class \emph{tool}, which is the sole category within this dataset. Examining the last column of Figure~\ref{fig:Worm-eating warbler}, we observe that the model's performance is compromised by both the non-specific nature of the word \emph{tool} and out-of-domain nature of the image. The generality of the term \emph{tool} sometimes leads to misinterpretation, with the model confusing it with elements of the camera interface itself. This ambiguity helps explain the substantial 35\% performance gap observed in this dataset.

Humans typically bridge this semantic gap by providing additional context \citep{pimentel2024speakerslexicalsemanticgaps}. However, in our experimental setup, this approach can be prohibitively expensive or unfeasible, as shown by the \emph{Worm-eating Warbler} case. 
While using the prompt ``bird'' could disambiguate this specific image, such generic prompts fail when working with datasets that include different bird species.
Visual Prompting offers a solution to this challenge by providing a simpler, less ambiguous method to fill this semantic gap, eliminating the need for elaborate textual descriptions or context-dependent prompts. In Appendix~\ref{sec:text_prompting_superiority} we conduct a similar analysis on the ambiguity of visual prompting.

Our considerations indicate that visual and text prompting are inherently complementary and that visual prompting offers a natural and readily available strategy to make up for the weaknesses of text prompting due to the identified ambiguities.

\section{PromptMatcher: combining text and visual prompts}
\label{sec:method}

\begin{figure}[t]
    \centering
    \includegraphics[width=1\textwidth]{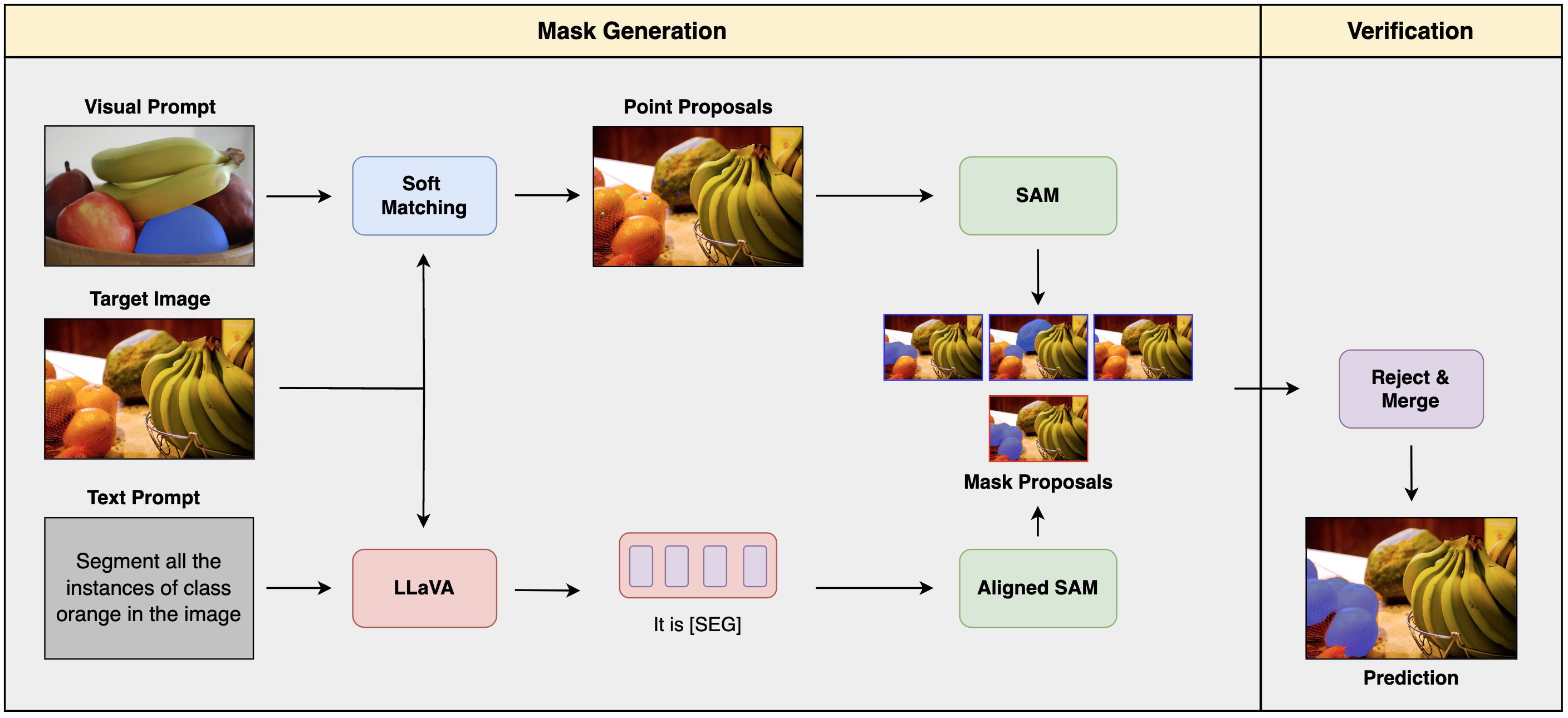} 
    \caption{PromptMatcher framework: The left section illustrates the mask generation process using visual and text prompts, while the right section shows the verification module which discards inaccurate predictions.}
    \label{fig:method graph} 
\end{figure}


Motivated by the complementary nature of text and visual prompts, we propose a framework that effectively integrates both, narrowing the gap between the baselines presented in Section~\ref{sec:ana} and the Oracle Ensemble+. Furthermore, drawing inspiration from LLM-Modulo frameworks outlined in \citep{kambhampati2024llmscantplanhelp}, particularly from the concept of employing critics/verifiers to enhance generative models' reasoning capabilities, in our context we propose to use SoftMatcher+ as an effective critic/verifier for LISA's predictions.
This verification module would be able to mitigate LISA's hallucinations, thereby enhancing overall accuracy.




We refer to our training-free framework as \emph{PromptMatcher}. As depicted in Figure~\ref{fig:method graph}, it employs SoftMatcher+ as both a critic and segmentation model, generating predictions using LISA for the text prompt branch and SoftMatcher+ for the visual prompt branch. Our framework consists of two steps: Mask generation and verification.
First, in the mask generation step, masks are generated by both the LISA and the SoftMatcher+ branch: the text prompt is processed by LISA's multi-modal LLaVA model, producing an output sequence with a specialized [SEG] token, which is then decoded into a segmentation mask by LISA's aligned SAM model.
Simultaneously, SoftMatcher+'s matching pipeline processes the visual prompt, generating multiple sets of point prompts representing potential object locations. The SAM mask-decoder uses these prompts to create unique output masks for each set of potential object locations.
Secondly, the masks from LISA and SoftMatcher+ are both passed to the verification step, where SoftMatcher+'s mask rejection pipeline is applied to masks produced by both branches (LISA and SoftMatcher+).
Masks that cannot be verified in their consistency with respect to the reference image are rejected and discarded. 
The verifier only allows plausible masks to pass, therefore playing the crucial role of a critic, reducing hallucinations originating from either branch.
Finally, the verified masks are combined by taking their union to form a single, comprehensive semantic segmentation output. Pseudocode detailing inner workings of the framework can be found in Appendix~\ref{section:pseudocode}

Regarding computation resources, PromptMatcher introduces minimal overhead on top of LISA. LISA itself is a 13B model (CLIP + LLaMA + SAM backbone + fine-tuned SAM head), while integrating SoftMatcher+ adds ~600k parameters in total (RadioViT-L ~300M + non-fine-tuned SAM head ~300K). Computationally, the PromptMatcher pipeline is equivalent to running SoftMatcher+ and LISA, with the additional rejection of LISA masks incurring negligible cost.

We present our results in Table~\ref{tab:performance_vision_language}, and refer to Table~\ref{table:performance_vision_language_full} in the Appendix~\ref{section:appendix_quantitative} for per-dataset results. Our combination of visual and text prompts significantly outperforms the vision-language SEEM baseline, which performs nearly the same as its vision-only version. We see that with our straightforward, training-free approach, it is possible to go beyond text-only or visual-only prompting and start to bridge the gap towards the Oracle Ensemble+. 
Notably, PromptMatcher surpasses Oracle Ensemble+ on two MESS datasets (DeepCrack and MHP v1), indicating synergies beyond simply selecting the better of two prompts. 
This superior performance can be attributed to the unique nature of the proposed framework.
As our approach leverages the complementary strengths of LISA and SoftMatcher+ to generate a more diverse set of predictions, when the outputs from the two models diverge, taking their union allows merging segments from different instances. This enables the models to combine their predicted masks rather than being limited to choose the output from one or the other, which is advantageous compared to an oracle-based selection.
Moreover, applying the mask rejection procedure from SoftMatcher+ to LISA masks helps to mitigate potential hallucinations from LISA by rejecting results that do not match with the reference mask.
The rejection of LISA masks capitalizes on the inherent text-vision knowledge distilled into the AM-RADIO representations, improving over vision-only backbones.



Our remarkably simple integration of TPs and VPs demonstrates the immediate benefit of combining the two modalities. We are convinced that there is untapped potential in such modular, training-free frameworks. We leave the exploration of more elaborate framework designs to future work, encouraging the research community's involvement in this effort.

\begin{table}[t]
\centering
\begin{tabular}{l|ccccc|c}
\toprule
& General & Earth & Medical & Engineering & Agriculture & Average \\
\midrule
SEEM & \textcolor{white}{0}9.7  & 17.0  & 20.5  & \textcolor{white}{0}7.3   & 22.5  & 15.4 \\
LISA & 57.0 & 47.7  & 31.7  & 12.8  & \textbf{64.0}  & 42.6 \\
SoftMatcher+    & 53.0 & 36.2  & 30.4  & 28.7  & 60.7  & 41.8 \\
PromptMatcher & \textbf{58.7} & \textbf{39.7}  & \textbf{35.1}  & \textbf{30.4} & 62.4 & \textbf{45.3}\\
\midrule
Oracle Ensemble+         & \textbf{67.3} & 51.8  & 46.2  & 32.5  & \textbf{71.4}  & 53.8 \\
Supervised      & 55.3 & \textbf{71.4}  & \textbf{82.6}  & \textbf{89.5}  & 62.8  & \textbf{72.3} \\
\bottomrule
\end{tabular}
\caption{Comparison of PromptMatcher's performance with i) SEEM using both visual and text prompts simultaneously ii) the top-performing text and visual prompt models, and iii) the Oracle Ensemble+ and the supervised baselines.}

\label{tab:performance_vision_language}
\end{table}

\subsection{PromptMatcher Ablation}
\label{sec:ablation}

In Table~\ref{table:matcher_ablation}, we present an ablation study evaluating various methods for integrating the LISA model with SoftMatcher+. Specifically, we compare the merging strategy of PromptMatcher, as described in Section~\ref{sec:method}, with alternative approaches, namely: (i) Probability Maps Merging, (ii) Cluster Merging, and (iii) TP/VP Selection. Moreover, we also analyze the impact of incorporating the LISA mask proposal when combined with the approaches described above.
\begin{itemize}
    \item \textbf{Probability Maps Merging:} This method combines the probability map generated by VP - which encodes similarity between the reference and target - with the probability map derived from TP. The latter is computed by applying a Softmax over the logits from LISA's aligned SAM decoder. The two probability maps are combined through addition, followed by renormalization of the outputs. The resulting map is then used for forward matching (to generate predictions) and backward matching (to reject masks). 

    \item \textbf{Cluster Merging:} This approach utilizes LISA's probability map - computed by applying a Softmax over the logits from LISA's aligned SAM decoder - to create clusters, from which points are sampled to generate new mask proposals. These proposals are then appended to those generated by SoftMatcher+. 

    \item \textbf{TP or VP Selection:} In this method, SoftMatcher+ and LISA operate independently, and the final mask is selected based on LISA's performance on the reference image. If LISA achieves an IoU greater than 20\%, the TP mask is selected; otherwise, the VP mask is used. 
    \item \textbf{LISA Mask:} All three methodologies described above can be enhanced by incorporating the LISA mask proposal alongside the one from SoftMatcher+ in the process of mask generation (analogous to Figure~\ref{fig:method graph}). Adding LISA's mask to the original SoftMatcher+ algorithm results in PromptMatcher. Note that adding the LISA mask to the VP in the \emph{TP or VP Selection} technique leads to a choice between LISA and PromptMatcher. Results of this ablation are highlighted in the second block of Table~\ref{table:matcher_ablation}.
\end{itemize}


A closer look at the results reveals distinct performance profiles. Merging-based techniques excel in specialized domains like medical and engineering datasets, leveraging visual information but underperforming in general tasks, lowering overall performance. In contrast, integrating the TP mask during generation boosts performance on general-purpose datasets by shifting the model's bias toward textual information, though at the cost of accuracy in technical domains. Surprisingly, the simplest approach is adding LISA's mask during generation, which offers the best trade-off, delivering balanced performance across datasets and achieving the highest average.

\begin{table}[t]
\centering
\begin{tabular}{l|ccccc|c}
\toprule
& General & Earth & Medical & Engineering & Agriculture & Average \\
\midrule
Probability Merging & 52.1 & 32.2 & \textbf{36.1} & 31.5 & 57.2 & 41.8 \\
Cluster Merging & 46.6  & 37.4 & 30.1 & \textbf{32.9} & 50.3 & 39.5 \\
TP or VP & 48.7 & 30.0 & 29.1 & 21.5 & 54.6 & 36.7 \\
\midrule
Probability Merging + LISA Mask & \textbf{59.3} & 40.7 & 33.9 & 23.0 & \textbf{63.1} & 44.0\\
Cluster Merging + LISA Mask & 54.0 & 37.2 & 31.5 & 27.7 & 58.0 & 41.7 \\

TP or VP + LISA Mask & 59.0 & \textbf{41.0} & 29.8 & 23.3 & 62.4 & 43.1 \\
\midrule
PromptMatcher (VP + LISA Mask) & 58.7 & 39.7  & 35.1  & 30.4 & 62.4 & \textbf{45.3}\\
\bottomrule

\end{tabular}
\label{table:matcher_ablation}
\caption{Ablation on PromptMatcher's construction. We compared different methods for integrating TP and VP, observing varying biases toward general or specialized datasets. Adding the TP mask directly to the VP predictions provided the most balanced approach, achieving the best average performance.}
\end{table}

\section{Related Work}

\textbf{Open-Vocabulary Segmentation Models} are able to perform segmentation across unlimited classes without relying on a fixed set of categories defined during training. These models often rely on CLIP-like text encoders to associate visual data with text descriptions. Specialized models like L-SEG \cite{li2022languagedrivensemanticsegmentation}, CAT-Seg \cite{cho2024catsegcostaggregationopenvocabulary}, and NACLIP \cite{hajimiri2024payattentionneighbourstrainingfree} are designed specifically to solve this task, while multi-modal models such as X-Decoder \cite{zou2022generalizeddecodingpixelimage} and SEEM \cite{zou2023segment} expand this capability by handling a different range of visual prompts. 

\textbf{Vision-Language Models} bridge the gap between visual perception and natural language understanding, excelling in tasks that require a combination of both, such as perception-language tasks and grounding tasks. These models are built using large language models (LLMs) integrated with vision encoders. With respect to perception-language tasks, VLMs perform tasks like image captioning, visual question answering, and region-level annotations. The LLaVA series \cite{liu2023llava, liu2023improvedllava, liu2024llavanext} has set benchmarks in this area by combining vision encoders like CLIP with LLMs, such as LLaMA \cite{touvron2023llamaopenefficientfoundation, touvron2023llama2openfoundation} or Vicuna \cite{vicuna2023}. InstructBLIP \cite{dai2023instructblipgeneralpurposevisionlanguagemodels} builds on the BLIP-2 \cite{li2023blip2bootstrappinglanguageimagepretraining} model with instruct tuning, and MM1 \cite{mckinzie2024mm1methodsanalysis} provides insights into crafting effective multimodal models. GPT-4V \cite{openai2024gpt4technicalreport} currently sets the highest standard in these perception-language tasks \cite{yang2023dawnlmmspreliminaryexplorations}. In grounding tasks, VLMs are able to handle phrase grounding and referring expression comprehension, detection, and segmentation. These tasks require identifying specific objects or regions based on text descriptions. Models like Florence-2 \cite{xiao2023florence2advancingunifiedrepresentation} predict segmentation coordinates in the form of text, while PALI-Gemma \cite{beyer2024paligemmaversatile3bvlm} uses a next-token prediction method encoding outputs to a fixed token dictionary, which is then decoded using a VQVAE \cite{oord2018neuraldiscreterepresentationlearning}. Other significant contributions include Kosmos-2 \cite{peng2023kosmos2groundingmultimodallarge}, which integrates coordinate tokens into the vocabulary for object detection, Ferret \cite{you2023ferretrefergroundgranularity}, which incorporates dense visual prompts, and Osprey \cite{yuan2024ospreypixelunderstandingvisual}, which adds further granularity to input prompts. While GPT-4V has shown impressive capabilities in many visual-language tasks, it has notable limitations in performing segmentation. Some VLMs incorporate specialized segmentation decoders, such as LISA \cite{lai2024lisareasoningsegmentationlarge}, which extends the LLaVA architecture incorporating SAM \cite{kirillov2023segment} to convert predicted tokens into segmentation masks. This hybrid approach has been refined by models like GLAMM \cite{rasheed2024glammpixelgroundinglarge}, which includes pixel-level visual prompting and supports multi-round conversations, and GSVA \cite{xia2024gsvageneralizedsegmentationmultimodal}, which enhances resilience to adversarial attacks. PixelLM \cite{ren2024pixellmpixelreasoninglarge} introduces a lightweight segmentation decoder, while SESAME \cite{wu2023seesaysegmentteaching} focuses on mitigating hallucination in segmentation tasks.

\textbf{Visual Prompting} involves providing visual cues to guide the model's understanding and segmentation of images. 
Early works such as \cite{bar2022visualpromptingimageinpainting}, focused on solving few-shot vision tasks by reconstructing the target via image inpainting of a grid-like input prompt. This concept was further developed in models like Painter \cite{wang2023imagesspeakimagesgeneralist} and SegGPT \cite{wang2023seggptsegmentingcontext}, which demonstrated the possibility of solving tasks like segmentation more effectively.
A significant leap forward came with the introduction of the Segment Anything Model (SAM) \cite{kirillov2023segment} and its follow-up \cite{ravi2024sam2segmentimages}, showing remarkable zero-shot capabilities in image segmentation tasks. These models, along with works like OMG-LLaVA \cite{zhang2024omgllavabridgingimagelevelobjectlevel}, focused on using visual prompts within the target image itself rather than relying on separate example images. 
Other notable works include DINOv \cite{li2023visualincontextprompting}, which extends visual prompting from SEEM, and T-Rex2 \cite{jiang2024trex2genericobjectdetection}, a concurrent study examining the role of modality in prompting for object detection. Matcher \cite{liu2024matchersegmentshotusing} brings a unique approach that enables zero-shot models like SAM to be prompted one-shot through feature matching. SoftMatcher \cite{matcher_vp} further expands on this concept by enhancing both simplicity and computation performance of the approach. Additionally, there has been growing research on optimizing information extraction from target images using pixel-level deformations. A seminal work in this direction is SoM \cite{yang2023setofmarkpromptingunleashesextraordinary}, which posited that providing visual clues to a VLM can significantly enhance its performance. This has sparked numerous follow-up studies, including ViP-LLaVA \cite{cai2024vipllavamakinglargemultimodal} that applies these concepts to models like LLaVA. The practical implications of these approaches are also being explored, such as the work \cite{he2024webvoyagerbuildingendtoendweb} in the context of web-based applications.


\section{Conclusion}

In this work, we introduced a benchmarking task designed to evaluate the performance of Vision-Language Models (VLMs) as semantic segmentation engines. Our results demonstrate that, despite the advancements, the latest VLMs still fall significantly short compared to custom models trained specifically on a given domain. This finding suggests that there is still room for progress in developing VLMs. We also showed that text prompting and visual prompting are complementary. By anticipating and selecting the most effective prompting modality, it is possible to achieve a notable 11\% IoU performance improvement. Building on this insight, we introduced a straightforward, training-free framework that leverages the complementary strengths of both text and visual prompting, with a key verification component responsible for rejecting incorrect segmentation masks. This framework sets a new state-of-the-art benchmark on the MESS dataset collection, achieving 45.5\% average IoU.
Our findings highlight the potential of using multiple prompt modalities to enhance the performance of VLMs without the need for additional training, bringing us closer to true foundation VLMs.
\clearpage

\bibliography{main}

\begin{thebibliography}{82}
\providecommand{\natexlab}[1]{#1}
\providecommand{\url}[1]{\texttt{#1}}
\expandafter\ifx\csname urlstyle\endcsname\relax
  \providecommand{\doi}[1]{doi: #1}\else
  \providecommand{\doi}{doi: \begingroup \urlstyle{rm}\Url}\fi

\bibitem[Bar et~al.(2022)Bar, Gandelsman, Darrell, Globerson, and
  Efros]{bar2022visualpromptingimageinpainting}
Amir Bar, Yossi Gandelsman, Trevor Darrell, Amir Globerson, and Alexei~A.
  Efros.
\newblock Visual prompting via image inpainting, 2022.
\newblock URL \url{https://arxiv.org/abs/2209.00647}.

\bibitem[Bashkirova et~al.(2022)Bashkirova, Abdelfattah, Zhu, Akl, Alladkani,
  Hu, Ablavsky, Calli, Bargal, and
  Saenko]{bashkirova2022zerowastedatasetdeformableobject}
Dina Bashkirova, Mohamed Abdelfattah, Ziliang Zhu, James Akl, Fadi Alladkani,
  Ping Hu, Vitaly Ablavsky, Berk Calli, Sarah~Adel Bargal, and Kate Saenko.
\newblock Zerowaste dataset: Towards deformable object segmentation in
  cluttered scenes, 2022.
\newblock URL \url{https://arxiv.org/abs/2106.02740}.

\bibitem[Beyer et~al.(2024)Beyer, Steiner, Pinto, Kolesnikov, Wang, Salz,
  Neumann, Alabdulmohsin, Tschannen, Bugliarello, Unterthiner, Keysers,
  Koppula, Liu, Grycner, Gritsenko, Houlsby, Kumar, Rong, Eisenschlos, Kabra,
  Bauer, Bošnjak, Chen, Minderer, Voigtlaender, Bica, Balazevic, Puigcerver,
  Papalampidi, Henaff, Xiong, Soricut, Harmsen, and
  Zhai]{beyer2024paligemmaversatile3bvlm}
Lucas Beyer, Andreas Steiner, André~Susano Pinto, Alexander Kolesnikov, Xiao
  Wang, Daniel Salz, Maxim Neumann, Ibrahim Alabdulmohsin, Michael Tschannen,
  Emanuele Bugliarello, Thomas Unterthiner, Daniel Keysers, Skanda Koppula,
  Fangyu Liu, Adam Grycner, Alexey Gritsenko, Neil Houlsby, Manoj Kumar, Keran
  Rong, Julian Eisenschlos, Rishabh Kabra, Matthias Bauer, Matko Bošnjak,
  Xi~Chen, Matthias Minderer, Paul Voigtlaender, Ioana Bica, Ivana Balazevic,
  Joan Puigcerver, Pinelopi Papalampidi, Olivier Henaff, Xi~Xiong, Radu
  Soricut, Jeremiah Harmsen, and Xiaohua Zhai.
\newblock Paligemma: A versatile 3b vlm for transfer, 2024.
\newblock URL \url{https://arxiv.org/abs/2407.07726}.

\bibitem[Bianchi \& Hebdon(2021)Bianchi and Hebdon]{Bianchi2021}
Eric Bianchi and Matthew Hebdon.
\newblock {Corrosion Condition State Semantic Segmentation Dataset}, 10 2021.
\newblock URL
  \url{https://data.lib.vt.edu/articles/dataset/Corrosion_Condition_State_Semantic_Segmentation_Dataset/16624663}.

\bibitem[Blumenstiel et~al.(2023)Blumenstiel, Jakubik, Kühne, and
  Vössing]{blumenstiel2023messmultidomainevaluationzeroshot}
Benedikt Blumenstiel, Johannes Jakubik, Hilde Kühne, and Michael Vössing.
\newblock What a mess: Multi-domain evaluation of zero-shot semantic
  segmentation, 2023.
\newblock URL \url{https://arxiv.org/abs/2306.15521}.

\bibitem[Cai et~al.(2024)Cai, Liu, Park, Mustikovela, Meyer, Chai, and
  Lee]{cai2024vipllavamakinglargemultimodal}
Mu~Cai, Haotian Liu, Dennis Park, Siva~Karthik Mustikovela, Gregory~P. Meyer,
  Yuning Chai, and Yong~Jae Lee.
\newblock Vip-llava: Making large multimodal models understand arbitrary visual
  prompts, 2024.
\newblock URL \url{https://arxiv.org/abs/2312.00784}.

\bibitem[Chiang et~al.(2023)Chiang, Li, Lin, Sheng, Wu, Zhang, Zheng, Zhuang,
  Zhuang, Gonzalez, Stoica, and Xing]{vicuna2023}
Wei-Lin Chiang, Zhuohan Li, Zi~Lin, Ying Sheng, Zhanghao Wu, Hao Zhang, Lianmin
  Zheng, Siyuan Zhuang, Yonghao Zhuang, Joseph~E. Gonzalez, Ion Stoica, and
  Eric~P. Xing.
\newblock Vicuna: An open-source chatbot impressing gpt-4 with 90\%* chatgpt
  quality, March 2023.
\newblock URL \url{https://lmsys.org/blog/2023-03-30-vicuna/}.

\bibitem[Cho et~al.(2024)Cho, Shin, Hong, Arnab, Seo, and
  Kim]{cho2024catsegcostaggregationopenvocabulary}
Seokju Cho, Heeseong Shin, Sunghwan Hong, Anurag Arnab, Paul~Hongsuck Seo, and
  Seungryong Kim.
\newblock Cat-seg: Cost aggregation for open-vocabulary semantic segmentation,
  2024.
\newblock URL \url{https://arxiv.org/abs/2303.11797}.

\bibitem[Church \& Patil(1982)Church and Patil]{church-patil-1982-coping}
Kenneth Church and Ramesh Patil.
\newblock Coping with syntactic ambiguity or how to put the block in the box on
  the table.
\newblock \emph{American Journal of Computational Linguistics}, 8\penalty0
  (3-4):\penalty0 139--149, 1982.
\newblock URL \url{https://aclanthology.org/J82-3004}.

\bibitem[Cohen et~al.(2022)Cohen, Newman, and
  Shamir]{cohen2022semanticsegmentationartpaintings}
Nadav Cohen, Yael Newman, and Ariel Shamir.
\newblock Semantic segmentation in art paintings, 2022.
\newblock URL \url{https://arxiv.org/abs/2203.03238}.

\bibitem[Dai et~al.(2023)Dai, Li, Li, Tiong, Zhao, Wang, Li, Fung, and
  Hoi]{dai2023instructblipgeneralpurposevisionlanguagemodels}
Wenliang Dai, Junnan Li, Dongxu Li, Anthony Meng~Huat Tiong, Junqi Zhao,
  Weisheng Wang, Boyang Li, Pascale Fung, and Steven Hoi.
\newblock Instructblip: Towards general-purpose vision-language models with
  instruction tuning, 2023.
\newblock URL \url{https://arxiv.org/abs/2305.06500}.

\bibitem[Erfani et~al.(2021)Erfani, Wu, Wu, Wang, and
  Goharian]{erfani2021atlantisbenchmarksemanticsegmentation}
Seyed Mohammad~Hassan Erfani, Zhenyao Wu, Xinyi Wu, Song Wang, and Erfan
  Goharian.
\newblock Atlantis: A benchmark for semantic segmentation of waterbody images,
  2021.
\newblock URL \url{https://arxiv.org/abs/2111.11567}.

\bibitem[et~al.(2023)]{touvron2023llama2openfoundation}
Hugo~Touvron et~al.
\newblock Llama 2: Open foundation and fine-tuned chat models, 2023.
\newblock URL \url{https://arxiv.org/abs/2307.09288}.

\bibitem[Everingham et~al.()Everingham, Van~Gool, Williams, Winn, and
  Zisserman]{pascal-voc-2012}
M.~Everingham, L.~Van~Gool, C.~K.~I. Williams, J.~Winn, and A.~Zisserman.
\newblock The {PASCAL} {V}isual {O}bject {C}lasses {C}hallenge 2012 {(VOC2012)}
  {R}esults.
\newblock
  http://www.pascal-network.org/challenges/VOC/voc2012/workshop/index.html.

\bibitem[Fraz et~al.(2012)Fraz, Remagnino, Hoppe, Uyyanonvara, Rudnicka, Owen,
  and Barman]{6224174}
Muhammad~Moazam Fraz, Paolo Remagnino, Andreas Hoppe, Bunyarit Uyyanonvara,
  Alicja~R. Rudnicka, Christopher~G. Owen, and Sarah~A. Barman.
\newblock An ensemble classification-based approach applied to retinal blood
  vessel segmentation.
\newblock \emph{IEEE Transactions on Biomedical Engineering}, 59\penalty0
  (9):\penalty0 2538--2548, 2012.
\newblock \doi{10.1109/TBME.2012.2205687}.

\bibitem[Frick et~al.(2024)Frick, Skura, Janicki, Assaf, Avogaro, Caraballo,
  Cinar, Ebouky, Giurgiu, Katsuki, Kluska, Malossi, Qiu, Sakai, Scheidegger,
  Simeski, Yang, Bartezzaghi, and Rigotti]{matcher_vp}
Thomas Frick, Cezary Skura, Filip Janicki, Roy Assaf, Niccolo Avogaro, Daniel
  Caraballo, Yagmur Cinar, Brown Ebouky, Ioana Giurgiu, Takayuki Katsuki, Piotr
  Kluska, A.~Cristiano~I. Malossi, Haoxiang Qiu, Tomoya Sakai, Florian
  Scheidegger, Andrej Simeski, Daniel Yang, Andrea Bartezzaghi, and Mattia
  Rigotti.
\newblock Probabilistic feature matching for fast scalable visual prompting,
  2024.

\bibitem[Gupta et~al.(2019)Gupta, Dollár, and
  Girshick]{gupta2019lvisdatasetlargevocabulary}
Agrim Gupta, Piotr Dollár, and Ross Girshick.
\newblock Lvis: A dataset for large vocabulary instance segmentation, 2019.
\newblock URL \url{https://arxiv.org/abs/1908.03195}.

\bibitem[Hajimiri et~al.(2024)Hajimiri, Ayed, and
  Dolz]{hajimiri2024payattentionneighbourstrainingfree}
Sina Hajimiri, Ismail~Ben Ayed, and Jose Dolz.
\newblock Pay attention to your neighbours: Training-free open-vocabulary
  semantic segmentation, 2024.
\newblock URL \url{https://arxiv.org/abs/2404.08181}.

\bibitem[Haug \& Ostermann(2015)Haug and Ostermann]{cwfid}
Sebastian Haug and J{\"o}rn Ostermann.
\newblock A crop/weed field image dataset for the evaluation of computer vision
  based precision agriculture tasks.
\newblock In Lourdes Agapito, Michael~M. Bronstein, and Carsten Rother (eds.),
  \emph{Computer Vision - ECCV 2014 Workshops}, pp.\  105--116, Cham, 2015.
  Springer International Publishing.
\newblock ISBN 978-3-319-16220-1.

\bibitem[He et~al.(2024)He, Yao, Ma, Yu, Dai, Zhang, Lan, and
  Yu]{he2024webvoyagerbuildingendtoendweb}
Hongliang He, Wenlin Yao, Kaixin Ma, Wenhao Yu, Yong Dai, Hongming Zhang,
  Zhenzhong Lan, and Dong Yu.
\newblock Webvoyager: Building an end-to-end web agent with large multimodal
  models, 2024.
\newblock URL \url{https://arxiv.org/abs/2401.13919}.

\bibitem[Islam et~al.(2020)Islam, Edge, Xiao, Luo, Mehtaz, Morse, Enan, and
  Sattar]{suim}
Md~Jahidul Islam, Chelsey Edge, Yuyang Xiao, Peigen Luo, Muntaqim Mehtaz,
  Christopher Morse, Sadman~Sakib Enan, and Junaed Sattar.
\newblock Semantic segmentation of underwater imagery: Dataset and benchmark.
\newblock In \emph{2020 IEEE/RSJ International Conference on Intelligent Robots
  and Systems (IROS)}, pp.\  1769--1776, 2020.
\newblock \doi{10.1109/IROS45743.2020.9340821}.

\bibitem[Jha et~al.(2021)Jha, Ali, Emanuelsen, Hicks, Thambawita, Garcia-Ceja,
  Riegler, de~Lange, Schmidt, Johansen, Johansen, and Halvorsen]{kvasir}
Debesh Jha, Sharib Ali, Krister Emanuelsen, Steven~A. Hicks, Vajira Thambawita,
  Enrique Garcia-Ceja, Michael~A. Riegler, Thomas de~Lange, Peter~T. Schmidt,
  H{\aa}vard~D. Johansen, Dag Johansen, and P{\aa}l Halvorsen.
\newblock Kvasir-instrument: Diagnostic and therapeutic tool segmentation
  dataset in gastrointestinal endoscopy.
\newblock In Jakub Loko{\v{c}}, Tom{\'a}{\v{s}} Skopal, Klaus Schoeffmann,
  Vasileios Mezaris, Xirong Li, Stefanos Vrochidis, and Ioannis Patras (eds.),
  \emph{MultiMedia Modeling}, pp.\  218--229, Cham, 2021. Springer
  International Publishing.
\newblock ISBN 978-3-030-67835-7.

\bibitem[Jiang et~al.(2024)Jiang, Li, Zeng, Ren, Liu, and
  Zhang]{jiang2024trex2genericobjectdetection}
Qing Jiang, Feng Li, Zhaoyang Zeng, Tianhe Ren, Shilong Liu, and Lei Zhang.
\newblock T-rex2: Towards generic object detection via text-visual prompt
  synergy, 2024.
\newblock URL \url{https://arxiv.org/abs/2403.14610}.

\bibitem[Kambhampati et~al.(2024)Kambhampati, Valmeekam, Guan, Verma, Stechly,
  Bhambri, Saldyt, and Murthy]{kambhampati2024llmscantplanhelp}
Subbarao Kambhampati, Karthik Valmeekam, Lin Guan, Mudit Verma, Kaya Stechly,
  Siddhant Bhambri, Lucas Saldyt, and Anil Murthy.
\newblock Llms can't plan, but can help planning in llm-modulo frameworks,
  2024.
\newblock URL \url{https://arxiv.org/abs/2402.01817}.

\bibitem[Kazemzadeh et~al.(2014)Kazemzadeh, Ordonez, andre Matten, and
  Berg]{Kazemzadeh2014ReferItGameRT}
Sahar Kazemzadeh, Vicente Ordonez, Marc andre Matten, and Tamara~L. Berg.
\newblock Referitgame: Referring to objects in photographs of natural scenes.
\newblock In \emph{Conference on Empirical Methods in Natural Language
  Processing}, 2014.
\newblock URL \url{https://api.semanticscholar.org/CorpusID:6308361}.

\bibitem[Kirillov et~al.(2023)Kirillov, Mintun, Ravi, Mao, Rolland, Gustafson,
  Xiao, Whitehead, Berg, Lo, Dollár, and Girshick]{kirillov2023segment}
Alexander Kirillov, Eric Mintun, Nikhila Ravi, Hanzi Mao, Chloe Rolland, Laura
  Gustafson, Tete Xiao, Spencer Whitehead, Alexander~C. Berg, Wan-Yen Lo, Piotr
  Dollár, and Ross Girshick.
\newblock Segment anything, 2023.
\newblock URL \url{https://arxiv.org/abs/2304.02643}.

\bibitem[Lai et~al.(2024)Lai, Tian, Chen, Li, Yuan, Liu, and
  Jia]{lai2024lisareasoningsegmentationlarge}
Xin Lai, Zhuotao Tian, Yukang Chen, Yanwei Li, Yuhui Yuan, Shu Liu, and Jiaya
  Jia.
\newblock Lisa: Reasoning segmentation via large language model, 2024.
\newblock URL \url{https://arxiv.org/abs/2308.00692}.

\bibitem[Li et~al.(2022)Li, Weinberger, Belongie, Koltun, and
  Ranftl]{li2022languagedrivensemanticsegmentation}
Boyi Li, Kilian~Q. Weinberger, Serge Belongie, Vladlen Koltun, and René
  Ranftl.
\newblock Language-driven semantic segmentation, 2022.
\newblock URL \url{https://arxiv.org/abs/2201.03546}.

\bibitem[Li et~al.(2023{\natexlab{a}})Li, Jiang, Zhang, Ren, Liu, Zou, Xu, Li,
  Li, Yang, Zhang, and Gao]{li2023visualincontextprompting}
Feng Li, Qing Jiang, Hao Zhang, Tianhe Ren, Shilong Liu, Xueyan Zou, Huaizhe
  Xu, Hongyang Li, Chunyuan Li, Jianwei Yang, Lei Zhang, and Jianfeng Gao.
\newblock Visual in-context prompting, 2023{\natexlab{a}}.
\newblock URL \url{https://arxiv.org/abs/2311.13601}.

\bibitem[Li et~al.(2018)Li, Zhao, Wei, Lang, Li, Sim, Yan, and
  Feng]{li2018multiplehumanparsingwild}
Jianshu Li, Jian Zhao, Yunchao Wei, Congyan Lang, Yidong Li, Terence Sim,
  Shuicheng Yan, and Jiashi Feng.
\newblock Multiple-human parsing in the wild, 2018.
\newblock URL \url{https://arxiv.org/abs/1705.07206}.

\bibitem[Li et~al.(2023{\natexlab{b}})Li, Li, Savarese, and
  Hoi]{li2023blip2bootstrappinglanguageimagepretraining}
Junnan Li, Dongxu Li, Silvio Savarese, and Steven Hoi.
\newblock Blip-2: Bootstrapping language-image pre-training with frozen image
  encoders and large language models, 2023{\natexlab{b}}.
\newblock URL \url{https://arxiv.org/abs/2301.12597}.

\bibitem[Lin et~al.(2015)Lin, Maire, Belongie, Bourdev, Girshick, Hays, Perona,
  Ramanan, Zitnick, and Dollár]{lin2015microsoftcococommonobjects}
Tsung-Yi Lin, Michael Maire, Serge Belongie, Lubomir Bourdev, Ross Girshick,
  James Hays, Pietro Perona, Deva Ramanan, C.~Lawrence Zitnick, and Piotr
  Dollár.
\newblock Microsoft coco: Common objects in context, 2015.
\newblock URL \url{https://arxiv.org/abs/1405.0312}.

\bibitem[Liu et~al.(2023{\natexlab{a}})Liu, Li, Li, and
  Lee]{liu2023improvedllava}
Haotian Liu, Chunyuan Li, Yuheng Li, and Yong~Jae Lee.
\newblock Improved baselines with visual instruction tuning,
  2023{\natexlab{a}}.

\bibitem[Liu et~al.(2023{\natexlab{b}})Liu, Li, Wu, and Lee]{liu2023llava}
Haotian Liu, Chunyuan Li, Qingyang Wu, and Yong~Jae Lee.
\newblock Visual instruction tuning, 2023{\natexlab{b}}.

\bibitem[Liu et~al.(2024{\natexlab{a}})Liu, Li, Li, Li, Zhang, Shen, and
  Lee]{liu2024llavanext}
Haotian Liu, Chunyuan Li, Yuheng Li, Bo~Li, Yuanhan Zhang, Sheng Shen, and
  Yong~Jae Lee.
\newblock Llava-next: Improved reasoning, ocr, and world knowledge, January
  2024{\natexlab{a}}.
\newblock URL \url{https://llava-vl.github.io/blog/2024-01-30-llava-next/}.

\bibitem[Liu et~al.(2019)Liu, Yao, Lu, Xie, and Li]{deepcrack}
Yahui Liu, Jian Yao, Xiaohu Lu, Renping Xie, and Li~Li.
\newblock Deepcrack: A deep hierarchical feature learning architecture for
  crack segmentation.
\newblock \emph{Neurocomput.}, 338\penalty0 (C):\penalty0 139–153, April
  2019.
\newblock ISSN 0925-2312.
\newblock \doi{10.1016/j.neucom.2019.01.036}.
\newblock URL \url{https://doi.org/10.1016/j.neucom.2019.01.036}.

\bibitem[Liu et~al.(2024{\natexlab{b}})Liu, Zhu, Li, Chen, Wang, and
  Shen]{liu2024matchersegmentshotusing}
Yang Liu, Muzhi Zhu, Hengtao Li, Hao Chen, Xinlong Wang, and Chunhua Shen.
\newblock Matcher: Segment anything with one shot using all-purpose feature
  matching, 2024{\natexlab{b}}.
\newblock URL \url{https://arxiv.org/abs/2305.13310}.

\bibitem[Lyu et~al.(2020)Lyu, Vosselman, Xia, Yilmaz, and
  Yang]{lyu2020uavidsemanticsegmentationdataset}
Ye~Lyu, George Vosselman, Guisong Xia, Alper Yilmaz, and Michael~Ying Yang.
\newblock Uavid: A semantic segmentation dataset for uav imagery, 2020.
\newblock URL \url{https://arxiv.org/abs/1810.10438}.

\bibitem[Mahbod et~al.(2021)Mahbod, Schaefer, Bancher, L{\"o}w, Dorffner,
  Ecker, and Ellinger]{Mahbod2021CryoNuSegAD}
Amirreza Mahbod, Gerald Schaefer, Benjamin Bancher, Christine L{\"o}w, Georg
  Dorffner, Rupert~C Ecker, and Isabella Ellinger.
\newblock Cryonuseg: A dataset for nuclei instance segmentation of
  cryosectioned h\&e-stained histological images.
\newblock \emph{Computers in biology and medicine}, 132:\penalty0 104349, 2021.

\bibitem[Manning \& Schutze(1999)Manning and Schutze]{Manning}
Christopher~D. Manning and Hinrich Schutze.
\newblock \emph{Foundations of statistical natural language processing}.
\newblock MIT Press, 1999.

\bibitem[Mao et~al.(2016)Mao, Huang, Toshev, Camburu, Yuille, and
  Murphy]{mao2016generationcomprehensionunambiguousobject}
Junhua Mao, Jonathan Huang, Alexander Toshev, Oana Camburu, Alan Yuille, and
  Kevin Murphy.
\newblock Generation and comprehension of unambiguous object descriptions,
  2016.
\newblock URL \url{https://arxiv.org/abs/1511.02283}.

\bibitem[Mateo-Garcia et~al.(2021)Mateo-Garcia, Veitch-Michaelis, Smith, Oprea,
  Schumann, Gal, Baydin, and Backes]{Mateo-Garcia2021}
Gonzalo Mateo-Garcia, Joshua Veitch-Michaelis, Lewis Smith, Silviu~Vlad Oprea,
  Guy Schumann, Yarin Gal, At{\i}l{\i}m~G{\"u}ne{\c{s}} Baydin, and Dietmar
  Backes.
\newblock Towards global flood mapping onboard low cost satellites with machine
  learning.
\newblock \emph{Scientific Reports}, 11\penalty0 (1):\penalty0 7249, Mar 2021.
\newblock ISSN 2045-2322.
\newblock \doi{10.1038/s41598-021-86650-z}.
\newblock URL \url{https://doi.org/10.1038/s41598-021-86650-z}.

\bibitem[McKinzie et~al.(2024)McKinzie, Gan, Fauconnier, Dodge, Zhang, Dufter,
  Shah, Du, Peng, Weers, Belyi, Zhang, Singh, Kang, Jain, Hè, Schwarzer,
  Gunter, Kong, Zhang, Wang, Wang, Du, Lei, Wiseman, Yin, Lee, Wang, Pang,
  Grasch, Toshev, and Yang]{mckinzie2024mm1methodsanalysis}
Brandon McKinzie, Zhe Gan, Jean-Philippe Fauconnier, Sam Dodge, Bowen Zhang,
  Philipp Dufter, Dhruti Shah, Xianzhi Du, Futang Peng, Floris Weers, Anton
  Belyi, Haotian Zhang, Karanjeet Singh, Doug Kang, Ankur Jain, Hongyu Hè, Max
  Schwarzer, Tom Gunter, Xiang Kong, Aonan Zhang, Jianyu Wang, Chong Wang, Nan
  Du, Tao Lei, Sam Wiseman, Guoli Yin, Mark Lee, Zirui Wang, Ruoming Pang,
  Peter Grasch, Alexander Toshev, and Yinfei Yang.
\newblock Mm1: Methods, analysis \& insights from multimodal llm pre-training,
  2024.
\newblock URL \url{https://arxiv.org/abs/2403.09611}.

\bibitem[OpenAI(2024)]{openai2024gpt4technicalreport}
OpenAI.
\newblock Gpt-4 technical report, 2024.
\newblock URL \url{https://arxiv.org/abs/2303.08774}.

\bibitem[Oquab et~al.(2024)Oquab, Darcet, Moutakanni, Vo, Szafraniec, Khalidov,
  Fernandez, Haziza, Massa, El-Nouby, Assran, Ballas, Galuba, Howes, Huang, Li,
  Misra, Rabbat, Sharma, Synnaeve, Xu, Jegou, Mairal, Labatut, Joulin, and
  Bojanowski]{oquab2024dinov2learningrobustvisual}
Maxime Oquab, Timothée Darcet, Théo Moutakanni, Huy Vo, Marc Szafraniec,
  Vasil Khalidov, Pierre Fernandez, Daniel Haziza, Francisco Massa, Alaaeldin
  El-Nouby, Mahmoud Assran, Nicolas Ballas, Wojciech Galuba, Russell Howes,
  Po-Yao Huang, Shang-Wen Li, Ishan Misra, Michael Rabbat, Vasu Sharma, Gabriel
  Synnaeve, Hu~Xu, Hervé Jegou, Julien Mairal, Patrick Labatut, Armand Joulin,
  and Piotr Bojanowski.
\newblock Dinov2: Learning robust visual features without supervision, 2024.
\newblock URL \url{https://arxiv.org/abs/2304.07193}.

\bibitem[Peng et~al.(2023)Peng, Wang, Dong, Hao, Huang, Ma, and
  Wei]{peng2023kosmos2groundingmultimodallarge}
Zhiliang Peng, Wenhui Wang, Li~Dong, Yaru Hao, Shaohan Huang, Shuming Ma, and
  Furu Wei.
\newblock Kosmos-2: Grounding multimodal large language models to the world,
  2023.
\newblock URL \url{https://arxiv.org/abs/2306.14824}.

\bibitem[Pimentel et~al.(2024)Pimentel, Maudslay, Blasi, and
  Cotterell]{pimentel2024speakerslexicalsemanticgaps}
Tiago Pimentel, Rowan~Hall Maudslay, Damián Blasi, and Ryan Cotterell.
\newblock Speakers fill lexical semantic gaps with context, 2024.
\newblock URL \url{https://arxiv.org/abs/2010.02172}.

\bibitem[Rahnemoonfar et~al.(2020)Rahnemoonfar, Chowdhury, Sarkar, Varshney,
  Yari, and Murphy]{rahnemoonfar2020floodnethighresolutionaerial}
Maryam Rahnemoonfar, Tashnim Chowdhury, Argho Sarkar, Debvrat Varshney, Masoud
  Yari, and Robin Murphy.
\newblock Floodnet: A high resolution aerial imagery dataset for post flood
  scene understanding, 2020.
\newblock URL \url{https://arxiv.org/abs/2012.02951}.

\bibitem[Ranzinger et~al.(2024)Ranzinger, Heinrich, Kautz, and
  Molchanov]{ranzinger2024amradioagglomerativevisionfoundation}
Mike Ranzinger, Greg Heinrich, Jan Kautz, and Pavlo Molchanov.
\newblock Am-radio: Agglomerative vision foundation model -- reduce all domains
  into one, 2024.
\newblock URL \url{https://arxiv.org/abs/2312.06709}.

\bibitem[Rasheed et~al.(2024)Rasheed, Maaz, Mullappilly, Shaker, Khan,
  Cholakkal, Anwer, Xing, Yang, and Khan]{rasheed2024glammpixelgroundinglarge}
Hanoona Rasheed, Muhammad Maaz, Sahal~Shaji Mullappilly, Abdelrahman Shaker,
  Salman Khan, Hisham Cholakkal, Rao~M. Anwer, Erix Xing, Ming-Hsuan Yang, and
  Fahad~S. Khan.
\newblock Glamm: Pixel grounding large multimodal model, 2024.
\newblock URL \url{https://arxiv.org/abs/2311.03356}.

\bibitem[Ravi et~al.(2024)Ravi, Gabeur, Hu, Hu, Ryali, Ma, Khedr, Rädle,
  Rolland, Gustafson, Mintun, Pan, Alwala, Carion, Wu, Girshick, Dollár, and
  Feichtenhofer]{ravi2024sam2segmentimages}
Nikhila Ravi, Valentin Gabeur, Yuan-Ting Hu, Ronghang Hu, Chaitanya Ryali,
  Tengyu Ma, Haitham Khedr, Roman Rädle, Chloe Rolland, Laura Gustafson, Eric
  Mintun, Junting Pan, Kalyan~Vasudev Alwala, Nicolas Carion, Chao-Yuan Wu,
  Ross Girshick, Piotr Dollár, and Christoph Feichtenhofer.
\newblock Sam 2: Segment anything in images and videos, 2024.
\newblock URL \url{https://arxiv.org/abs/2408.00714}.

\bibitem[Ren et~al.(2024)Ren, Huang, Wei, Zhao, Fu, Feng, and
  Jin]{ren2024pixellmpixelreasoninglarge}
Zhongwei Ren, Zhicheng Huang, Yunchao Wei, Yao Zhao, Dongmei Fu, Jiashi Feng,
  and Xiaojie Jin.
\newblock Pixellm: Pixel reasoning with large multimodal model, 2024.
\newblock URL \url{https://arxiv.org/abs/2312.02228}.

\bibitem[Riemer(1949)]{Riemer_1950}
Svend Riemer.
\newblock \emph{Human Behavior and the Principle of Least Effort}.
\newblock Cambridge: Addison Wesley Press, 1949.

\bibitem[Rottensteiner et~al.(2012)Rottensteiner, Sohn, Jung, Gerke, Baillard,
  Bénitez, and Breitkopf]{isprs}
Franz Rottensteiner, Gunho Sohn, Jaewook Jung, Markus Gerke, Caroline Baillard,
  Sébastien Bénitez, and U~Breitkopf.
\newblock The isprs benchmark on urban object classification and 3d building
  reconstruction.
\newblock \emph{ISPRS Annals of Photogrammetry, Remote Sensing and Spatial
  Information Sciences}, I-3, 07 2012.
\newblock \doi{10.5194/isprsannals-I-3-293-2012}.

\bibitem[Sakaridis et~al.(2019)Sakaridis, Dai, and
  Gool]{sakaridis2019guidedcurriculummodeladaptation}
Christos Sakaridis, Dengxin Dai, and Luc~Van Gool.
\newblock Guided curriculum model adaptation and uncertainty-aware evaluation
  for semantic nighttime image segmentation, 2019.
\newblock URL \url{https://arxiv.org/abs/1901.05946}.

\bibitem[Seibold et~al.(2022)Seibold, Reiß, Sarfraz, Fink, Mayer, Sellner,
  Kim, Maier-Hein, Kleesiek, and
  Stiefelhagen]{seibold2022detailedannotationschestxrays}
Constantin Seibold, Simon Reiß, Saquib Sarfraz, Matthias~A. Fink, Victoria
  Mayer, Jan Sellner, Moon~Sung Kim, Klaus~H. Maier-Hein, Jens Kleesiek, and
  Rainer Stiefelhagen.
\newblock Detailed annotations of chest x-rays via ct projection for report
  understanding, 2022.
\newblock URL \url{https://arxiv.org/abs/2210.03416}.

\bibitem[Shivakumar et~al.(2019)Shivakumar, Rodrigues, Zhou, Miller, Kumar, and
  Taylor]{shivakumar2019pst900rgbthermalcalibrationdataset}
Shreyas~S. Shivakumar, Neil Rodrigues, Alex Zhou, Ian~D. Miller, Vijay Kumar,
  and Camillo~J. Taylor.
\newblock Pst900: Rgb-thermal calibration, dataset and segmentation network,
  2019.
\newblock URL \url{https://arxiv.org/abs/1909.10980}.

\bibitem[Touvron et~al.(2023)Touvron, Lavril, Izacard, Martinet, Lachaux,
  Lacroix, Rozière, Goyal, Hambro, Azhar, Rodriguez, Joulin, Grave, and
  Lample]{touvron2023llamaopenefficientfoundation}
Hugo Touvron, Thibaut Lavril, Gautier Izacard, Xavier Martinet, Marie-Anne
  Lachaux, Timothée Lacroix, Baptiste Rozière, Naman Goyal, Eric Hambro,
  Faisal Azhar, Aurelien Rodriguez, Armand Joulin, Edouard Grave, and Guillaume
  Lample.
\newblock Llama: Open and efficient foundation language models, 2023.
\newblock URL \url{https://arxiv.org/abs/2302.13971}.

\bibitem[van~den Oord et~al.(2018)van~den Oord, Vinyals, and
  Kavukcuoglu]{oord2018neuraldiscreterepresentationlearning}
Aaron van~den Oord, Oriol Vinyals, and Koray Kavukcuoglu.
\newblock Neural discrete representation learning, 2018.
\newblock URL \url{https://arxiv.org/abs/1711.00937}.

\bibitem[Wah et~al.(2011)Wah, Branson, Welinder, Perona, and
  Belongie]{wah_branson_welinder_perona_belongie_2011}
Catherine Wah, Steve Branson, Peter Welinder, Pietro Perona, and Serge
  Belongie.
\newblock The caltech-ucsd birds-200-2011 dataset, Jul 2011.

\bibitem[Wang et~al.(2023{\natexlab{a}})Wang, Wang, Cao, Shen, and
  Huang]{wang2023imagesspeakimagesgeneralist}
Xinlong Wang, Wen Wang, Yue Cao, Chunhua Shen, and Tiejun Huang.
\newblock Images speak in images: A generalist painter for in-context visual
  learning, 2023{\natexlab{a}}.
\newblock URL \url{https://arxiv.org/abs/2212.02499}.

\bibitem[Wang et~al.(2023{\natexlab{b}})Wang, Zhang, Cao, Wang, Shen, and
  Huang]{wang2023seggptsegmentingcontext}
Xinlong Wang, Xiaosong Zhang, Yue Cao, Wen Wang, Chunhua Shen, and Tiejun
  Huang.
\newblock Seggpt: Segmenting everything in context, 2023{\natexlab{b}}.
\newblock URL \url{https://arxiv.org/abs/2304.03284}.

\bibitem[Wang et~al.(2023{\natexlab{c}})Wang, Wei, Schuurmans, Le, Chi, Narang,
  Chowdhery, and Zhou]{wang2023selfconsistencyimproveschainthought}
Xuezhi Wang, Jason Wei, Dale Schuurmans, Quoc Le, Ed~Chi, Sharan Narang,
  Aakanksha Chowdhery, and Denny Zhou.
\newblock Self-consistency improves chain of thought reasoning in language
  models, 2023{\natexlab{c}}.
\newblock URL \url{https://arxiv.org/abs/2203.11171}.

\bibitem[Wasow(2015)]{Wasow2015-WASAAI}
Thomas Wasow.
\newblock Ambiguity avoidance is overrated.
\newblock In Susanne Winkler (ed.), \emph{Ambiguity: Language and
  Communication}, pp.\  29--48. De Gruyter, 2015.

\bibitem[Wei et~al.(2023)Wei, Wang, Schuurmans, Bosma, Ichter, Xia, Chi, Le,
  and Zhou]{wei2023chainofthoughtpromptingelicitsreasoning}
Jason Wei, Xuezhi Wang, Dale Schuurmans, Maarten Bosma, Brian Ichter, Fei Xia,
  Ed~Chi, Quoc Le, and Denny Zhou.
\newblock Chain-of-thought prompting elicits reasoning in large language
  models, 2023.
\newblock URL \url{https://arxiv.org/abs/2201.11903}.

\bibitem[Wu et~al.(2023)Wu, Biamby, Chan, Dunlap, Gupta, Wang, Gonzalez, and
  Darrell]{wu2023seesaysegmentteaching}
Tsung-Han Wu, Giscard Biamby, David Chan, Lisa Dunlap, Ritwik Gupta, Xudong
  Wang, Joseph~E. Gonzalez, and Trevor Darrell.
\newblock See, say, and segment: Teaching lmms to overcome false premises,
  2023.
\newblock URL \url{https://arxiv.org/abs/2312.08366}.

\bibitem[Wu et~al.(2021)Wu, Fu, Liu, Lim, Hoi, and
  Sun]{wu2021largescalebenchmarkfoodimage}
Xiongwei Wu, Xin Fu, Ying Liu, Ee-Peng Lim, Steven C.~H. Hoi, and Qianru Sun.
\newblock A large-scale benchmark for food image segmentation, 2021.
\newblock URL \url{https://arxiv.org/abs/2105.05409}.

\bibitem[Xia et~al.(2024)Xia, Han, Han, Pan, Song, and
  Huang]{xia2024gsvageneralizedsegmentationmultimodal}
Zhuofan Xia, Dongchen Han, Yizeng Han, Xuran Pan, Shiji Song, and Gao Huang.
\newblock Gsva: Generalized segmentation via multimodal large language models,
  2024.
\newblock URL \url{https://arxiv.org/abs/2312.10103}.

\bibitem[Xiao et~al.(2023)Xiao, Wu, Xu, Dai, Hu, Lu, Zeng, Liu, and
  Yuan]{xiao2023florence2advancingunifiedrepresentation}
Bin Xiao, Haiping Wu, Weijian Xu, Xiyang Dai, Houdong Hu, Yumao Lu, Michael
  Zeng, Ce~Liu, and Lu~Yuan.
\newblock Florence-2: Advancing a unified representation for a variety of
  vision tasks, 2023.
\newblock URL \url{https://arxiv.org/abs/2311.06242}.

\bibitem[Yang et~al.(2023{\natexlab{a}})Yang, Zhang, Li, Zou, Li, and
  Gao]{yang2023setofmarkpromptingunleashesextraordinary}
Jianwei Yang, Hao Zhang, Feng Li, Xueyan Zou, Chunyuan Li, and Jianfeng Gao.
\newblock Set-of-mark prompting unleashes extraordinary visual grounding in
  gpt-4v, 2023{\natexlab{a}}.
\newblock URL \url{https://arxiv.org/abs/2310.11441}.

\bibitem[Yang et~al.(2023{\natexlab{b}})Yang, Li, Lin, Wang, Lin, Liu, and
  Wang]{yang2023dawnlmmspreliminaryexplorations}
Zhengyuan Yang, Linjie Li, Kevin Lin, Jianfeng Wang, Chung-Ching Lin, Zicheng
  Liu, and Lijuan Wang.
\newblock The dawn of lmms: Preliminary explorations with gpt-4v(ision),
  2023{\natexlab{b}}.
\newblock URL \url{https://arxiv.org/abs/2309.17421}.

\bibitem[Yao et~al.(2023)Yao, Yu, Zhao, Shafran, Griffiths, Cao, and
  Narasimhan]{yao2023treethoughtsdeliberateproblem}
Shunyu Yao, Dian Yu, Jeffrey Zhao, Izhak Shafran, Thomas~L. Griffiths, Yuan
  Cao, and Karthik Narasimhan.
\newblock Tree of thoughts: Deliberate problem solving with large language
  models, 2023.
\newblock URL \url{https://arxiv.org/abs/2305.10601}.

\bibitem[You et~al.(2023)You, Zhang, Gan, Du, Zhang, Wang, Cao, Chang, and
  Yang]{you2023ferretrefergroundgranularity}
Haoxuan You, Haotian Zhang, Zhe Gan, Xianzhi Du, Bowen Zhang, Zirui Wang,
  Liangliang Cao, Shih-Fu Chang, and Yinfei Yang.
\newblock Ferret: Refer and ground anything anywhere at any granularity, 2023.
\newblock URL \url{https://arxiv.org/abs/2310.07704}.

\bibitem[Yu et~al.(2020)Yu, Chen, Wang, Xian, Chen, Liu, Madhavan, and
  Darrell]{yu2020bdd100kdiversedrivingdataset}
Fisher Yu, Haofeng Chen, Xin Wang, Wenqi Xian, Yingying Chen, Fangchen Liu,
  Vashisht Madhavan, and Trevor Darrell.
\newblock Bdd100k: A diverse driving dataset for heterogeneous multitask
  learning, 2020.
\newblock URL \url{https://arxiv.org/abs/1805.04687}.

\bibitem[Yuan et~al.(2024)Yuan, Li, Liu, Tang, Luo, Qin, Zhang, and
  Zhu]{yuan2024ospreypixelunderstandingvisual}
Yuqian Yuan, Wentong Li, Jian Liu, Dongqi Tang, Xinjie Luo, Chi Qin, Lei Zhang,
  and Jianke Zhu.
\newblock Osprey: Pixel understanding with visual instruction tuning, 2024.
\newblock URL \url{https://arxiv.org/abs/2312.10032}.

\bibitem[Zamir et~al.(2019)Zamir, Arora, Gupta, Khan, Sun, Khan, Zhu, Shao,
  Xia, and Bai]{zamir2019isaidlargescaledatasetinstance}
Syed~Waqas Zamir, Aditya Arora, Akshita Gupta, Salman Khan, Guolei Sun,
  Fahad~Shahbaz Khan, Fan Zhu, Ling Shao, Gui-Song Xia, and Xiang Bai.
\newblock isaid: A large-scale dataset for instance segmentation in aerial
  images, 2019.
\newblock URL \url{https://arxiv.org/abs/1905.12886}.

\bibitem[Zhang et~al.(2024{\natexlab{a}})Zhang, Li, Fei, Yuan, Wu, Ji, Loy, and
  Yan]{zhang2024omgllavabridgingimagelevelobjectlevel}
Tao Zhang, Xiangtai Li, Hao Fei, Haobo Yuan, Shengqiong Wu, Shunping Ji,
  Chen~Change Loy, and Shuicheng Yan.
\newblock Omg-llava: Bridging image-level, object-level, pixel-level reasoning
  and understanding, 2024{\natexlab{a}}.
\newblock URL \url{https://arxiv.org/abs/2406.19389}.

\bibitem[Zhang et~al.(2024{\natexlab{b}})Zhang, Zhang, Li, Zhao, Karypis, and
  Smola]{zhang2024multimodalchainofthoughtreasoninglanguage}
Zhuosheng Zhang, Aston Zhang, Mu~Li, Hai Zhao, George Karypis, and Alex Smola.
\newblock Multimodal chain-of-thought reasoning in language models,
  2024{\natexlab{b}}.
\newblock URL \url{https://arxiv.org/abs/2302.00923}.

\bibitem[Zhou et~al.(2018)Zhou, Zhao, Puig, Xiao, Fidler, Barriuso, and
  Torralba]{zhou2018semanticunderstandingscenesade20k}
Bolei Zhou, Hang Zhao, Xavier Puig, Tete Xiao, Sanja Fidler, Adela Barriuso,
  and Antonio Torralba.
\newblock Semantic understanding of scenes through the ade20k dataset, 2018.
\newblock URL \url{https://arxiv.org/abs/1608.05442}.

\bibitem[Zhou et~al.(2023)Zhou, Schärli, Hou, Wei, Scales, Wang, Schuurmans,
  Cui, Bousquet, Le, and Chi]{zhou2023leasttomostpromptingenablescomplex}
Denny Zhou, Nathanael Schärli, Le~Hou, Jason Wei, Nathan Scales, Xuezhi Wang,
  Dale Schuurmans, Claire Cui, Olivier Bousquet, Quoc Le, and Ed~Chi.
\newblock Least-to-most prompting enables complex reasoning in large language
  models, 2023.
\newblock URL \url{https://arxiv.org/abs/2205.10625}.

\bibitem[Zou et~al.(2022)Zou, Dou, Yang, Gan, Li, Li, Dai, Behl, Wang, Yuan,
  Peng, Wang, Lee, and Gao]{zou2022generalizeddecodingpixelimage}
Xueyan Zou, Zi-Yi Dou, Jianwei Yang, Zhe Gan, Linjie Li, Chunyuan Li, Xiyang
  Dai, Harkirat Behl, Jianfeng Wang, Lu~Yuan, Nanyun Peng, Lijuan Wang,
  Yong~Jae Lee, and Jianfeng Gao.
\newblock Generalized decoding for pixel, image, and language, 2022.
\newblock URL \url{https://arxiv.org/abs/2212.11270}.

\bibitem[Zou et~al.(2023)Zou, Yang, Zhang, Li, Li, Wang, Wang, Gao, and
  Lee]{zou2023segment}
Xueyan Zou, Jianwei Yang, Hao Zhang, Feng Li, Linjie Li, Jianfeng Wang, Lijuan
  Wang, Jianfeng Gao, and Yong~Jae Lee.
\newblock Segment everything everywhere all at once, 2023.
\newblock URL \url{https://arxiv.org/abs/2304.06718}.

\end{thebibliography}
\bibliographystyle{tmlr}

\clearpage

\appendix
\section{Appendix}
\subsection{MESS dataset composition}
\label{sec:appendix_mess}

MESS Dataset integrates 22 datasets selected for their unique challenges, grouped into General, Earth, Medical, Engineering, and Agriculture domains. It evaluates model performance on out-of-distribution and adversarial examples, featuring visually complex medical images like those in Kvasir-Inst., and granular subclass divisions of common categories as seen in FoodSeg103 \cite{wu2021largescalebenchmarkfoodimage} and Caltech-UCSD Birds \cite{wah_branson_welinder_perona_belongie_2011} datasets. Table \ref{tab:mess-groups} displays the dataset grouping breakdown.

\begin{table}[t]
\centering
\small
  \begin{tabular}{l|l}
  \toprule
    General & ATLANTIS~\citep{erfani2021atlantisbenchmarksemanticsegmentation}, BDD100K~\citep{yu2020bdd100kdiversedrivingdataset},\\
    & Dark~Zurich~\citep{sakaridis2019guidedcurriculummodeladaptation}, DRAM~\citep{cohen2022semanticsegmentationartpaintings},\\
    & FoodSeg103~\citep{wu2021largescalebenchmarkfoodimage}, MHPv1~\citep{li2018multiplehumanparsingwild}\\ \midrule
    Earth &    FloodNet~\citep{rahnemoonfar2020floodnethighresolutionaerial}, iSAID~\citep{zamir2019isaidlargescaledatasetinstance},\\
    & ISPRS~Potsdam~\citep{isprs}, UAVid~\citep{lyu2020uavidsemanticsegmentationdataset},\\
    & WorldFloods~\citep{Mateo-Garcia2021}\\ \midrule
    Medical &
    CHASE~DB1~\citep{6224174},
    CryoNuSeg~\citep{Mahbod2021CryoNuSegAD},\\
    &
    Kvasir-Inst.~\citep{kvasir},
    PAXRay-4~\citep{seibold2022detailedannotationschestxrays}\\ \midrule
    Engineering &
    Corrosion~CS~\citep{Bianchi2021}, DeepCrack~\citep{deepcrack},\\
    & PST900~\citep{shivakumar2019pst900rgbthermalcalibrationdataset},
    ZeroWaste-f~\citep{bashkirova2022zerowastedatasetdeformableobject}\\ \midrule
    Agriculture &
    CUB-200~\citep{wah_branson_welinder_perona_belongie_2011},
    CWFID~\citep{cwfid},\\
    & SUIM~\citep{suim}\\ \bottomrule
  \end{tabular}
  \caption{Grouping of datasets in the MESS collection~\citep{blumenstiel2023messmultidomainevaluationzeroshot}.}
  \label{tab:mess-groups}
\end{table}

\subsection{Extended qualitative analysis}
\label{section:appendix_qualitative}

\begin{figure}[t]
    \centering
    \includegraphics[width=0.9\textwidth]{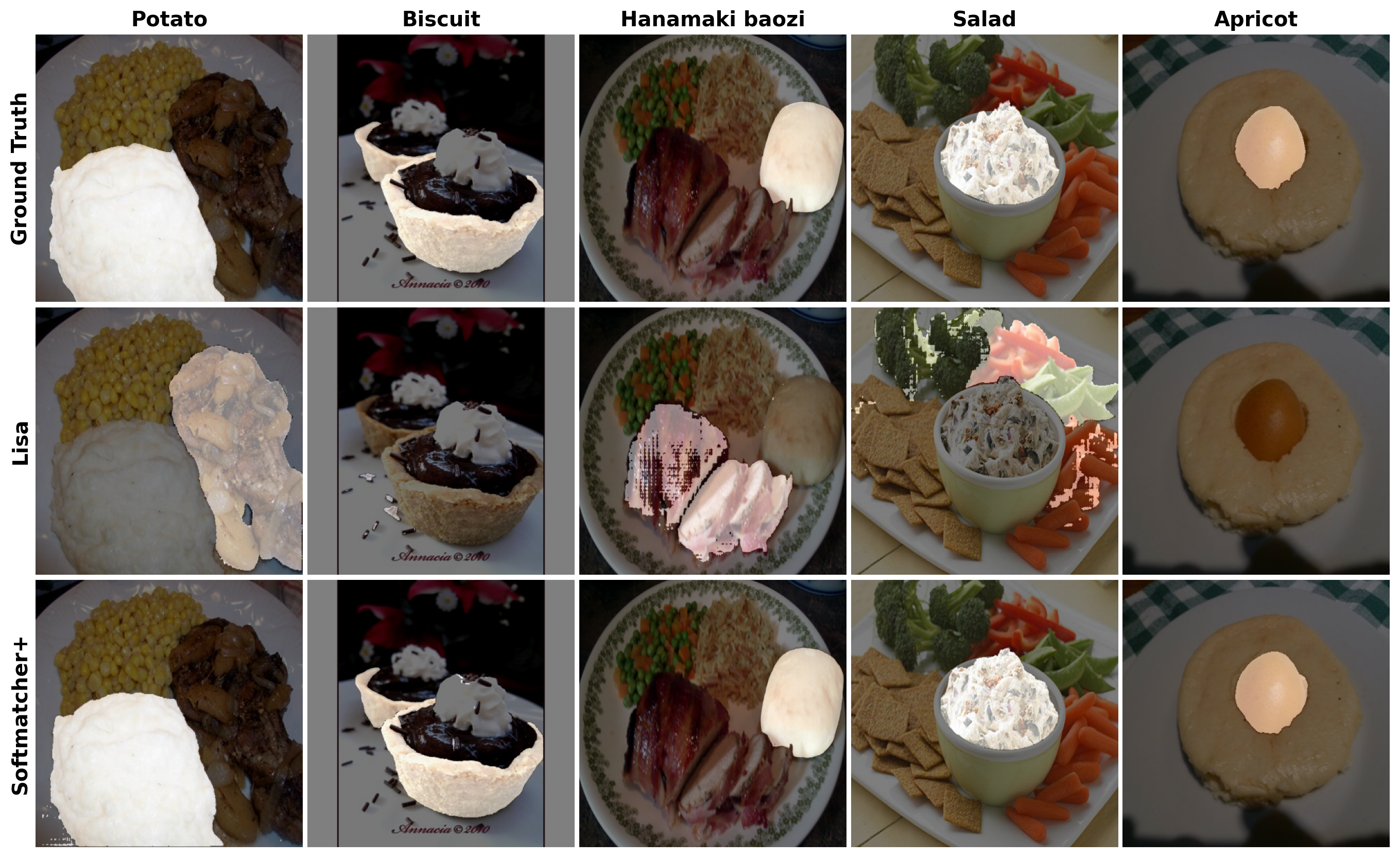} 
    \caption{Qualitative examples selected from the most challenging classes of FoodSeg103.}
    \label{fig:foodseg_images} 
\end{figure}

Figure \ref{fig:foodseg_images} showcases additional examples where LISA encounters difficulties with certain classes in FoodSeg103. These images are selected from specific categories that proved challenging for the model.
In the first image, LISA struggles to identify \emph{mashed potato}, possibly due to its transformed state from the raw ingredient. The second image presents a biscuit-based cake, where the model incorrectly focuses on crumbs rather than recognizing the entire structure as \emph{biscuit}.
The \emph{Hanamaki Baozi} example represents an out-of-domain concept, similar to the previously discussed Worm-eating Warbler case, highlighting the model's limitations with unfamiliar items. In the salad image, LISA misinterprets individual vegetables as the salad itself rather than recognizing the complete dish.
Lastly, an adversarial example shows an apricot that visually resembles an egg, causing the model to fail in producing any output. This highlights LISA's vulnerability to visual similarities that deviate from expected appearances within a class.
These examples illustrate the ongoing challenges in visual recognition tasks, particularly when dealing with transformed ingredients, culturally specific items, composite dishes, and visually ambiguous subjects.

\begin{figure}[t]
    \centering
    \includegraphics[width=0.9\textwidth]{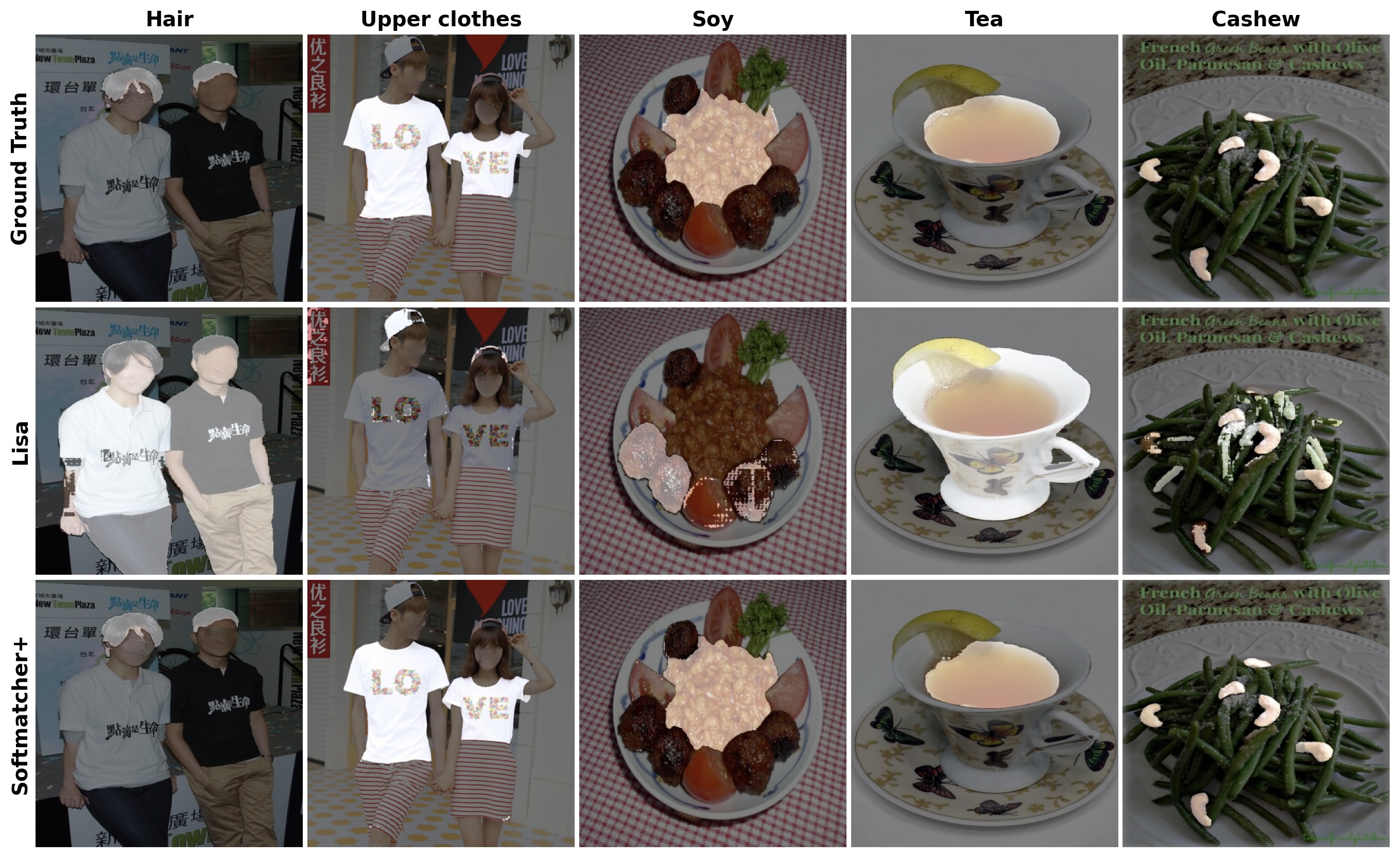} 
    \caption{Qualitative analysis on examples of challenging classes for Text Prompting.}
    \label{fig:anon_res} 
\end{figure}

Figure \ref{fig:anon_res} presents additional visual examples of the top 10 classes that posed challenges for LISA.
The \emph{hair} class consistently proves problematic, with LISA often predicting the entire person instead of isolating the hair. For \emph{upper clothes}, the model's misinterpretation can be attributed to linguistic ambiguity; in this instance, LISA incorrectly identified headwear as upper clothing despite being more accurately classified as an accessory.
In the \emph{soy} example, LISA fails to segment the soybean, instead erroneously detecting meatballs. The \emph{tea} image shows the model including the cup in its segmentation rather than isolating the liquid alone.
The final example demonstrates partial success, with LISA correctly identifying some cashews. However, it also exhibits a strong bias towards detecting non-relevant vegetables, leading to over-segmentation.

\newpage

\subsection{Text Prompting Superiority}
\label{sec:text_prompting_superiority}

\begin{figure}[t]
    \centering
    \includegraphics[width=0.9\textwidth]{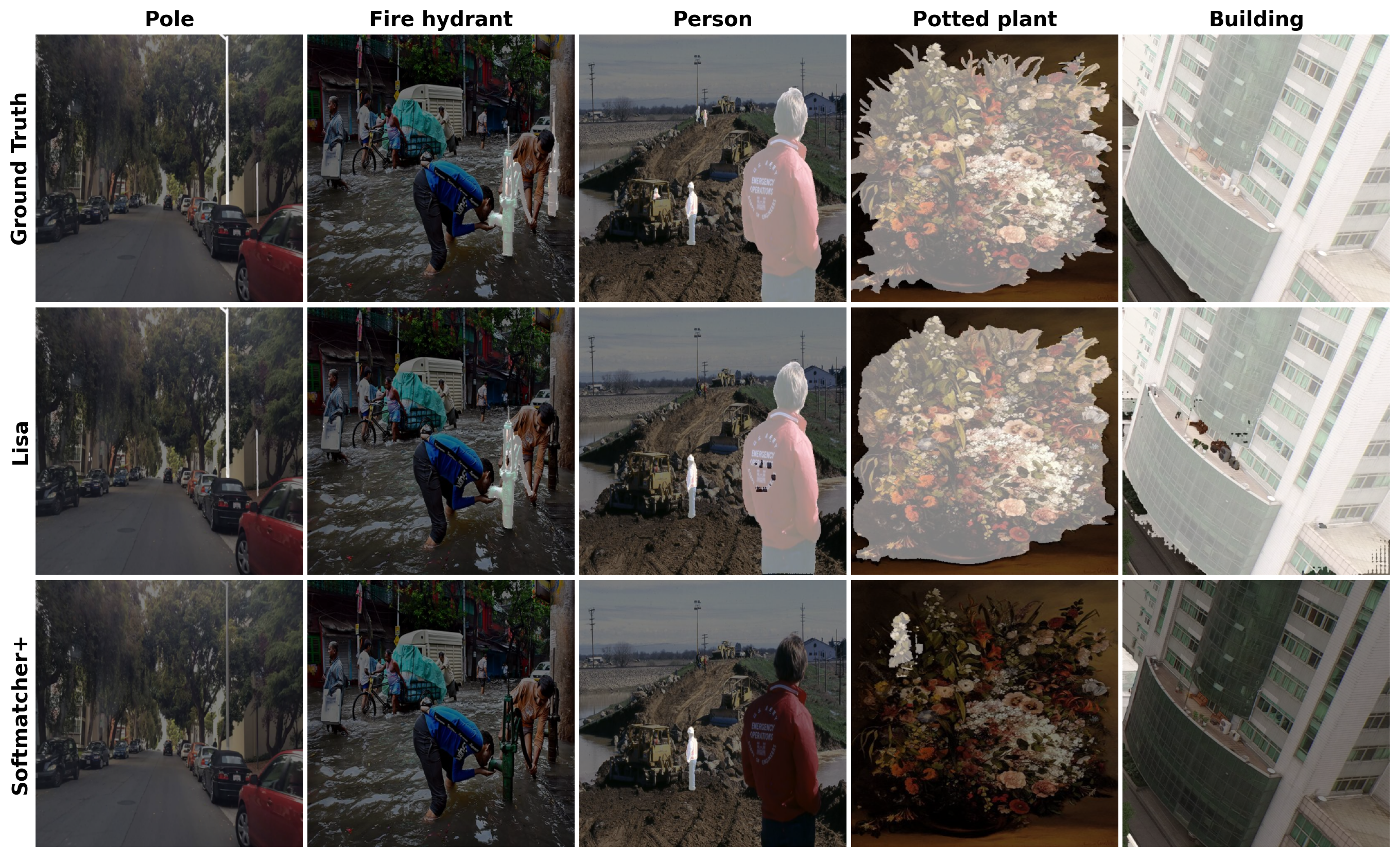} 
    \caption{Qualitative analysis of examples where text prompting excels. Classes like Potted Plant and Building can vary significantly in appearance, making it challenging for SoftMatcher+ to generate accurate predictions.}
    \label{fig:grid_tp} 
\end{figure}

We perform a mirrored analysis of Section~\ref{sec:text_ambiguity} to better understand when LISA outperforms SoftMatcher+. Specifically, we sort the per-class IoU results and report the top 10 classes where TP surpasses VP in Table~\ref{tab:negative_performance_difference}. Additionally, in Figure~\ref{fig:grid_tp}, we present the images with the largest difference per class for the top five classes.

Results indicate that LISA performs best in classes aligned with its training data. In fact, 9 out of 10 classes on the list appear in its training datasets (e.g., Pole, Building → ADE20K; Fire Hydrant → RefCOCOg; Person, Potted Plant, Boat → COCO). This suggests that the evaluation of these classes is largely in-domain. The alignment between test classes and training data further explains why LISA outperforms specialized models trained in-domain on ``General'' datasets,  as pointed out in Section~\ref{sec:text_prompting}

On the other hand, we attribute VP’s failure in these classes primarily to the broad internal variation within each category. Classes like \textit{building} and \textit{boat} cover a vast range of visual diversity. For instance, \textit{boat} includes everything from freighters to rowboats, which in order to be solved a prompt optimization would be needed, in a specular way to what would be done in language. For instance, while the general term ``bird'' might work for identifying a \textit{worm-eating warbler}, a more specific image prompt of a freighter would be more effective than using a general image of a \textit{boat} for identifying a freighter.

\begin{table}[t]
\centering
\begin{tabular}{l|ccc}
\toprule
Class name & IoU TP & IoU VP & Difference \\
\midrule
Pole \textit{(BDD100K)} & 41.71 & \textcolor{white}{0}7.64 & 34.07 \\
Fire Hydrant & 33.50 & \textcolor{white}{0}0.00 & 33.50 \\
Person \textit{(ATLANTIS)} & 58.33 & 25.77 & 32.56 \\
Potted Plant & 72.37 & 40.20 & 32.17 \\
Building \textit{(UAVid)} & 66.64 & 34.89 & 31.75 \\
White Pelican & 94.32 & 64.48 & 29.84 \\
Person \textit{(DRAM)} & 78.82 & 49.04 & 29.78 \\
Pole \textit{(ATLANTIS)} & 33.58 & \textcolor{white}{0}4.92 & 28.66 \\
Building \textit{(Dark Zurich)} & 59.75 & 31.49 & 28.26 \\
Boat & 50.98 & 23.50 & 27.48 \\
\bottomrule
\end{tabular}
\caption{Top 10 classes with the highest IoU difference between text- and visual-prompted models. Results show that LISA outperforms SoftMatcher+ on classes encountered during training.}
\label{tab:negative_performance_difference}
\end{table}

\subsection{In-Domain Performance}
\label{sec:indomain}

In this section, we explain why we intentionally avoid the traditional in-domain model performance evaluation. In Table~\ref{tab:coco}, we show how our proposed method compares to LISA, SoftMatcher+, and traditional few-shot pipelines on standard few-shot semantic segmentation datasets like Pascal-5i and COCO-20i. LISA alone significantly outperforms the chosen baselines from the FSS literature and SoftMatcher+, as it was trained in-domain on the validation classes such as COCO, refCOCO and ADE20k among others. The proposed PromptMatcher, which strives to balance LISA and SoftMatcher+ doesn't reach LISA's performance levels, primarily due to the performance of the visual prompting branch, which performs significantly worse on these types of datasets than LISA.

The results support our claim that VLMs trained on massive internet-scale datasets with domains similar to (or the same as, in the case of COCO) the traditional datasets, perform exceptionally well in-domain. However, this strong in-domain performance does not translate to technical out-of-domain performance, which more closely mirrors real-world use cases.
As a result, performance on traditional datasets is not a reliable indicator of the few-shot performance of the underlying model.

\begin{table}[t]
\centering
\begin{tabular}{l|cc}
\toprule
Method & COCO-20$^i$ & Pascal-5$^i$ \\
\midrule
Painter &	32.80 &	64.50 \\
Seggpt	& 56.10	& \textbf{83.20} \\
PAGMA-Net \textit{(CLIP-RN101)}	& \textbf{59.40}	& 77.60 \\
HMNet	& 52.10	& 70.40 \\
\midrule
LISA	& \textbf{72.23}	& \textbf{80.97} \\
SoftMatcher+	& 55.12	& 67.98 \\
\midrule
PromptMatcher	& 59.07	& 77.13 \\
\bottomrule
\end{tabular}
\caption{In-domain performance on FSS Datasets.}
\label{tab:coco}
\end{table}

\subsection{Extended quantitative analysis}
\label{section:appendix_quantitative}

Tables \ref{table:performance_text_only_full} and \ref{table:performance_vision_only_full} present comprehensive results for text prompted and vision-only models on MESS datasets, respectively. Table \ref{table:task_full} shows oracle results, while Table \ref{table:performance_vision_language_full} displays TP-VP framework outcomes.

\subsection{PromptMatcher Pseudocode}
\label{section:pseudocode}

Algoritm~\ref{alg:mask_pipeline} showcases PromptMatcher pseudocode.

\begin{table}[b]
\centering
\scriptsize
\begin{tabular}{llccccccc}
\toprule
& Dataset & SEEM txt	& CAT-Seg &  Florence &  PALI-Gem	& NACLIP & LISA & Supervised \\
\midrule
\multirow{6}{*}{\rotatebox{90}{General}}
& ATLANTIS  & 48.4 & 30.5 & 14.4 & 46.8 & 46.79 & 63.9 & 45.1 \\
& BDD100K  & 32.6 & 30.6 & \textcolor{white}{0}4.5 & 25.9 &  27.54 & 78.0 & 82.3 \\
& Dark Zurich  & 33.1 & 45.8 & 11.4 & 21.8 & 34.37 & 41.1 & 44.8 \\
& DRAM  & 60.4 & 33.6 & 29.3 & 58.6 & 50.05 & 78.6 & 42.2 \\
& FoodSeg103  & 31.0 & 30.0 & 18.1 & 51.3 & 37.81 & 60.6 & 53.2 \\
& MHP v1 & 10.0 & 33.1 & \textcolor{white}{0}6.5 & \textcolor{white}{0}7.6 & 19.77 & 19.8 & 63.9 \\
\midrule
\multirow{5}{*}{\rotatebox{90}{Earth}}
& FloodNet & 59.6 & \textcolor{white}{0}9.2 & 28.6 & 62.5 & 66.35 & 72.9 & 84.6 \\
& iSAID  & \textcolor{white}{0}9.5 & 66.5 & \textcolor{white}{0}4.1 & \textcolor{white}{0}4.3 & \textcolor{white}{0}9.80 & 31.3 & 45.7 \\
& ISPRS Potsdam $\!\!\!\!$ & 40.7 & 53.9 & 11.0 & 23.9 & 39.36 & 41.0 & 74.0 \\
& UAVid & 57.5 & 39.0 & 11.5 & 34.7 & 56.44 & 59.8 & 87.2 \\
& WorldFloods  & 16.9 & 16.1 & 14.4 & 20.3 & 33.94 & 33.4 & 65.3 \\
\midrule
\multirow{4}{*}{\rotatebox{90}{Medical}}
& CHASE DB1  & \textcolor{white}{0}9.8 & 49.9 & \textcolor{white}{0}9.1 & \textcolor{white}{0}8.9 & 10.05 & 16.7 & 92.7 \\
& CryoNuSeg & 24.1 & 39.8 & \textcolor{white}{0}6.7 & 24.2 & 24.77 & 31.9 & 82.2 \\
& Kvasir-Inst.  & 28.6 & 51.4 & 10.2 & 44.9 & 12.97 & 23.2 & 87.6 \\
& PAXRay-4 & 53.1 & 42.0 & 26.7 & 35.7 & 43.11 & 54.9 & 67.8 \\
\midrule
\multirow{4}{*}{\rotatebox{90}{Engin.}}
& Corrosion CS  & 11.1 & 25.0 & \textcolor{white}{0}7.7 & \textcolor{white}{0}8.8 & \textcolor{white}{0}4.47 & 13.8 & 97.1 \\
& DeepCrack  & \textcolor{white}{0}4.2 & 35.1 & \textcolor{white}{0}5.5 & \textcolor{white}{0}4.5 & \textcolor{white}{0}4.78 & \textcolor{white}{0}6.8 & 73.5 \\
& PST900  & 14.3 & 79.4 & \textcolor{white}{0}6.3 & \textcolor{white}{0}2.9 & \textcolor{white}{0}3.87 & 12.1 & 93.7 \\
& ZeroWaste-f  & 26.2 & 54.5 & \textcolor{white}{0}9.8 & 12.9 & 13.93 & 18.5 & 93.8 \\
\midrule
\multirow{3}{*}{\rotatebox{90}{Agri.}}
& CUB-200  & 89.0 & 31.4 & \textcolor{white}{0}0.0 & 68.2 & 14.36 & 88.1 & 85.9 \\
& CWFID & 13.7 & 25.3 & \textcolor{white}{0}4.2 & \textcolor{white}{0}7.0 & 11.79 & 36.6 & 52.5 \\
& SUIM & 31.0 & 16.9 & 18.7 & 44.9 & 40.86 & 67.2 & 49.9 \\
\bottomrule
\end{tabular}
\caption{Per dataset performance of text prompted methods}
\label{table:performance_text_only_full}
\end{table}

\begin{table}[t]
\centering
\scriptsize
\begin{tabular}{llccccc}
\toprule
& Dataset & SEEM vis & DINOv & VP & SoftMatcher+ & Supervised \\
\midrule
\multirow{6}{*}{\rotatebox{90}{General}}
& ATLANTIS  & 15.8 & 52.8 & 45.0 & 51.4 & 45.1 \\
& BDD100K  & \textcolor{white}{0}7.2 & 37.8 & 53.1 & 58.5 & 82.3 \\
& Dark Zurich  & \textcolor{white}{0}4.0 & 22.6 & 45.4 & 47.7 & 44.8 \\
& DRAM  & 13.4 & 73.6 & 55.9 & 62.9 & 42.2 \\
& FoodSeg103  & 11.8 & 28.3 & 54.0 & 60.5 & 53.2 \\
& MHP v1 & \textcolor{white}{0}5.6 & \textcolor{white}{0}9.5 & 34.6 & 36.7 & 63.9 \\
\midrule
\multirow{5}{*}{\rotatebox{90}{Earth}}
& FloodNet & 41.6 & 59.9 & 56.7 & 57.4 & 84.6 \\
& iSAID  & \textcolor{white}{0}2.2 & \textcolor{white}{0}4.3 & 22.8 & 26.7 & 45.7 \\
& ISPRS Potsdam  & 13.0 & 24.2 & 41.2 & 41.4 & 74.0 \\
& UAVid & 15.5 & 34.5 & 32.7 & 35.7 & 87.2 \\
& WorldFloods  & 11.9 & 17.3 & 16.4 & 20.0 & 65.3 \\
\midrule
\multirow{4}{*}{\rotatebox{90}{Medical}}
& CHASE DB1  & 10.4 & \textcolor{white}{0}9.6 & \textcolor{white}{0}0.0 & \textcolor{white}{0}0.0 & 92.7 \\
& CryoNuSeg & 26.8 & 24.0 & 21.2 & 24.5 & 82.2 \\
& Kvasir-Inst.  & \textcolor{white}{0}6.5 & 24.4 & 65.7 & 58.0 & 87.6 \\
& PAXRay-4 & 38.1 & 39.0 & 39.0 & 39.1 & 67.8 \\
\midrule
\multirow{4}{*}{\rotatebox{90}{Engin.}}
& Corrosion CS  & \textcolor{white}{0}9.3 & 10.1 & \textcolor{white}{0}7.2 & 14.8 & 97.1 \\
& DeepCrack  & \textcolor{white}{0}3.6 & \textcolor{white}{0}4.5 & 30.7 & 39.3 & 73.5 \\
& PST900  & \textcolor{white}{0}4.5 & \textcolor{white}{0}4.8 & 16.4 & 38.9 & 93.7 \\
& ZeroWaste-f  & 10.4 & 13.9 & 21.0 & 21.9 & 93.8 \\
\midrule
\multirow{3}{*}{\rotatebox{90}{Agri.}}
& CUB-200  & 20.7 & 92.2 & 85.4 & 87.0 & 85.9 \\
& CWFID & 17.5 & 33.5 & 41.5 & 41.0 & 52.5 \\
& SUIM & 26.9 & 51.4 & 52.5 & 54.1 & 49.9 \\
\bottomrule
\end{tabular}
\caption{Per dataset performance of visual prompted methods}
\label{table:performance_vision_only_full}
\end{table}

\begin{table}[t]
\centering
\scriptsize
\begin{tabular}{llccccc}
\toprule
& Dataset & SoftMatcher+ & LISA & Oracle & Oracle+ & Supervised \\
\midrule
\multirow{6}{*}{\rotatebox{90}{General}}
& ATLANTIS & 51.4 & 63.9 & 63.9 & 68.9 & 45.1 \\
& BDD100K & 58.5 & 78.0 & 78.0 & 79.2 & 82.3 \\
& Dark Zurich & 47.7 & 41.1 & 47.7 & 55.0 & 44.8 \\
& DRAM & 62.9 & 78.6 & 78.6 & 81.3 & 42.2 \\
& FoodSeg103 & 60.5 & 60.6 & 60.6 & 74.0 & 53.2 \\
& MHP v1 & 36.7 & 19.8 & 36.7 & 45.3 & 63.9 \\
\midrule
\multirow{5}{*}{\rotatebox{90}{Earth}}
& FloodNet & 57.4 & 72.9 & 72.9 & 74.8 & 84.6 \\
& iSAID & 26.7 & 31.3 & 31.3 & 35.4 & 45.7 \\
& ISPRS Potsdam & 41.4 & 41.0 & 41.4 & 50.2 & 74.0 \\
& UAVid & 35.7 & 59.8 & 59.8 & 65.0 & 87.2 \\
& WorldFloods & 20.0 & 33.4 & 33.4 & 33.4 & 65.3 \\
\midrule
\multirow{4}{*}{\rotatebox{90}{Medical}}
& CHASE DB1 & \textcolor{white}{0}0.0 & 16.7 & 16.7 & 16.7 & 92.7 \\
& CryoNuSeg & 24.5 & 31.9 & 31.9 & 34.5 & 82.2 \\
& Kvasir-Inst. & 58.0 & 23.2 & 58.0 & 72.0 & 87.6 \\
& PAXRay-4 & 39.1 & 54.9 & 54.9 & 61.7 & 67.8 \\
\midrule
\multirow{4}{*}{\rotatebox{90}{Engin.}}
& Corrosion CS & 14.8 & 13.8 & 14.8 & 17.6 & 97.1 \\
& DeepCrack & 39.3 & \textcolor{white}{0}6.8 & 39.3 & 42.2 & 73.5 \\
& PST900 & 38.9 & 12.1 & 38.7 & 39.7 & 93.7 \\
& ZeroWaste-f & 21.9 & 18.5 & 21.9 & 30.5 & 93.8 \\
\midrule
\multirow{3}{*}{\rotatebox{90}{Agri.}}
& CUB-200 & 87.0 & 88.1 & 88.1 & 90.5 & 85.9 \\
& CWFID & 41.0 & 36.6 & 41.0 & 48.4 & 52.5 \\
& SUIM & 54.1 & 67.2 & 67.2 & 75.2 & 49.9 \\
\bottomrule
\end{tabular}
\caption{Per dataset performance of Oracle ensembling baselines.}
\label{table:task_full}
\end{table}

\begin{table}[t]
\centering
\scriptsize
\begin{tabular}{llcccccc}
\toprule
 &Dataset & SEEM & LISA & SoftMatcher+ & PromptMatcher & Oracle+ & Supervised \\
\midrule
\multirow{6}{*}{\rotatebox{90}{General}}
& ATLANTIS  & 15.8 & 63.9 & 51.4 & 55.7 & 68.9 & 45.1 \\
& BDD100K  & \textcolor{white}{0}6.9 & 78.0 & 58.5 & 67.3 & 79.2 & 82.3 \\
& Dark Zurich  & \textcolor{white}{0}4.3 & 41.1 & 47.7 & 51.7 & 55.0 & 44.8 \\
& DRAM  & 13.5 & 78.6 & 62.9 & 69.7 & 81.3 & 42.2 \\
& FoodSeg103  & 12.0 & 60.6 & 60.7 & 61.9 & 74.0 & 53.2 \\
& MHP v1 & \textcolor{white}{0}5.8 & 19.8 & 36.7 & 46.2 & 45.3 & 63.9 \\
\midrule
\multirow{5}{*}{\rotatebox{90}{Earth}}
& FloodNet & 40.7 & 72.9 & 57.4 & 61.4 & 74.8 & 84.6 \\
& iSAID  & \textcolor{white}{0}2.3 & 31.3 & 26.7 & 24.3 & 35.4 & 45.7 \\
& ISPRS Potsdam & 13.1 & 41.0 & 41.4 & 45.9 & 50.2 & 74.0 \\
& UAVid & 14.9 & 59.8 & 35.7 & 52.4 & 65.0 & 87.2 \\
& WorldFloods  & 14.2 & 33.4 & 20.0 & 14.7 & 33.4 & 65.3 \\
\midrule
\multirow{4}{*}{\rotatebox{90}{Medical}}
& CHASE DB1  & 10.4 & 16.7 & \textcolor{white}{0}0.0 & \textcolor{white}{0}0.0 & 16.7 & 92.7 \\
& CryoNuSeg & 27.1 & 31.9 & 24.5 & 24.1 & 34.5 & 82.2 \\
& Kvasir-Inst.  & \textcolor{white}{0}6.4 & 23.2 & 58.0 & 60.8 & 72.0 & 87.6 \\
& PAXRay-4 & 38.1 & 54.9 & 39.1 & 55.5 & 61.7 & 67.8 \\
\midrule
\multirow{4}{*}{\rotatebox{90}{Engin.}}
& Corrosion CS  & 10.4 & 13.8 & 14.8 & 15.2 & 17.6 & 97.1 \\
& DeepCrack  & \textcolor{white}{0}3.8 & \textcolor{white}{0}6.8 & 39.3 & 42.6 & 42.2 & 73.5 \\
& PST900  & \textcolor{white}{0}4.9 & 12.1 & 38.9 & 39.3 & 39.9 & 93.7 \\
& ZeroWaste-f  & 10.1 & 18.5 & 21.9 & 24.6 & 30.5 & 93.8 \\
\midrule
\multirow{3}{*}{\rotatebox{90}{Agri.}}
& CUB-200  & 21.1 & 88.1 & 87.0 & 88.9 & 90.5 & 85.9 \\
& CWFID & 17.5 & 36.6 & 41.0 & 38.4 & 48.4 & 52.5 \\
& SUIM & 28.8 & 67.2 & 54.1 & 59.8 & 75.2 & 49.9 \\
\bottomrule
\end{tabular}
\caption{Per dataset performance of visual-text prompted methods}
\label{table:performance_vision_language_full}
\end{table}

\begin{algorithm}
\caption{\textbf{PromptMatcher}}
\label{alg:mask_pipeline}
\SetAlgoLined
\KwIn{$\mathit{reference\_image}$, $\mathit{reference\_mask}$, $\mathit{reference\_text}$,$\mathit{target\_image}$}
\KwOut{$\mathit{final\_mask}$}
\BlankLine
\BlankLine

$\mathit{ref\_feats} \leftarrow \mathrm{extract\_features}(\mathit{reference\_image})$\tcp*{Extract Features}
$\mathit{targ\_feats} \leftarrow \mathrm{extract\_features}(\mathit{target\_image})$\;
$\mathit{targ\_sam\_feats} \leftarrow \mathrm{extract\_SAM\_features}(\mathit{target\_image})$\;
\BlankLine
\BlankLine

$\mathit{probability\_map} \leftarrow \mathrm{match\_features}(\mathit{ref\_feats}, \mathit{reference\_mask}, \mathit{targ\_feats})$\tcp*{SoftMatcher+}
$\mathit{prompt\_points} \leftarrow \mathrm{sample\_and\_cluster}(\mathit{probability\_map})$\;
$\mathit{softmatcher\_masks} \leftarrow \mathrm{SAM\_mask\_decoder}(\mathit{prompt\_points}, \mathit{target\_sam\_feats})$\;
\BlankLine
\BlankLine

$\mathit{lisa\_SEG\_token} \leftarrow \mathrm{LISA\_VLM}(\mathit{target\_image}, \mathit{reference\_text})$\tcp*{LISA}
$\mathit{lisa\_mask} \leftarrow \mathrm{LISA\_mask\_decoder}(\mathit{target\_sam
\_feats}, \mathit{LISA\_SEG\_token})$ \;
\BlankLine
\BlankLine

$\mathit{mask\_proposals} \leftarrow \mathit{lisa\_mask} + \mathit{mask\_proposals}$ \tcp*{Merge masks}
\BlankLine
\BlankLine

$\mathit{masks} \leftarrow \mathrm{reject\_masks}(\mathit{mask\_proposals})$ \tcp*{Reject and merge masks}
$\mathit{segmentation\_mask} \leftarrow \mathrm{merge\_masks}(\mathit{masks})$\;
\BlankLine
\BlankLine

\Return $\mathit{segmentation\_mask}$
\end{algorithm}

\end{document}